\documentclass[10pt,twocolumn,letterpaper]{article}
\pdfoutput=1
\usepackage[pagenumbers]{cvpr} % To force page numbers, e.g. for an arXiv version

\usepackage{color}

\newcommand{\redfont}[1]{\textcolor{red}{#1}}
\newcommand{\bluefont}[1]{\textcolor{blue}{#1}}

\usepackage{multirow}
\usepackage[dvipsnames]{xcolor}
\definecolor{cvprblue}{rgb}{0.21,0.49,0.74}
\usepackage[pagebackref,breaklinks,colorlinks,citecolor=cvprblue]{hyperref}

\def\paperID{5003} % *** Enter the Paper ID here
\def\confName{CVPR}
\def\confYear{2024}

\title{MyPortrait: Morphable Prior-Guided Personalized Portrait Generation}
\author{Bo Ding$^{1,2}$ ~~~Zhenfeng Fan$^{1,2}$ ~~~Shuang Yang$^{1,2}$ ~~~Shihong Xia$^{1,2}$\\
$^1$Institute of Computing Technology, Chinese Academy of Sciences\\
$^2$University of Chinese Academy of Sciences\\
{\tt\small \{dingbo21s, fanzhenfeng, yangshuang21b, xsh\}@ict.ac.cn}
}

\begin{document}

% *****************************************************
% maniuscipt
% *****************************************************
\maketitle
\begin{abstract}
Generating realistic talking faces is an interesting and long-standing topic in the field of computer vision. Although significant progress has been made, it is still challenging to generate high-quality dynamic faces with personalized details. This is mainly due to the inability of the general model to represent personalized details and the generalization problem to unseen controllable parameters. In this work, we propose Myportrait, a simple, general, and flexible framework for neural portrait generation. We incorporate personalized prior in a monocular video and morphable prior in 3D face morphable space for generating personalized details under novel controllable parameters. Our proposed framework supports both video-driven and audio-driven face animation given a monocular video of a single person. Distinguished by whether the test data is sent to training or not, our method provides a real-time online version and a high-quality offline version. Comprehensive experiments in various metrics demonstrate the superior performance of our method over the state-of-the-art methods. The code will be publicly available.

\end{abstract}    
\section{Introduction}
\label{sec:intro}

The goal of talking face generation is to generate realistic talking faces whose motion is consistent with the control signals (commonly face video or speech audio), which is a long-standing topic with promising applications in digital human generation, film production, and video dubbing \cite{zhen2023human,dhere2020review,sha2023deep,zhan2023multimodal}. Recently, significant progress has been made in existing works on neural portrait generation with the advancement of deep learning technologies. Some sophisticated methods are able to generate fake videos which are hardly distinguishable from the real ones. 

Existing works on neural portrait generation can be categorized into 2D-based and 3D-based methods. On the one hand, 2D-based methods \cite{siarohin2019first,hong2022depth,wang2022latent,oorloff2023robust} directly process the image without resorting to facial semantic information (\eg pose and expression). These methods are user-friendly and easy to operate. However, 2D-based methods commonly neglect out-of-plane facial motions and are suboptimal in dealing with pose and expression variations. On the other hand, 3D-based methods \cite{ren2021pirenderer, grassal2022neural, xu2023latentavatar,wang2023robust,wang2023efficient,medin2022most} generate facial movements from 3D semantic information, \eg pose and expression parameters from 3D morphable models (3DMM) \cite{blanz1999morphable,booth20163d,paysan20093d,li2017learning}. These methods achieve flexible control of facial motion but commonly require complex network structures and expensive training costs. Some state-of-the-art methods \cite{yin2022styleheat,sun2023next3d} require a universal facial generator in addition to an input monocular video to recover missing pixels for unseen poses and expressions. As a side effect, a universal generator, \eg StyleGAN \cite{karras2019style,karras2020analyzing,karras2021alias} and EG3D \cite{chan2022efficient}, cannot match personalized characters and generally leads to the degradation of image quality, \eg missing high-frequency details in the results.

\begin{figure}[t]
    \centering
    \includegraphics[width = 0.47\textwidth]{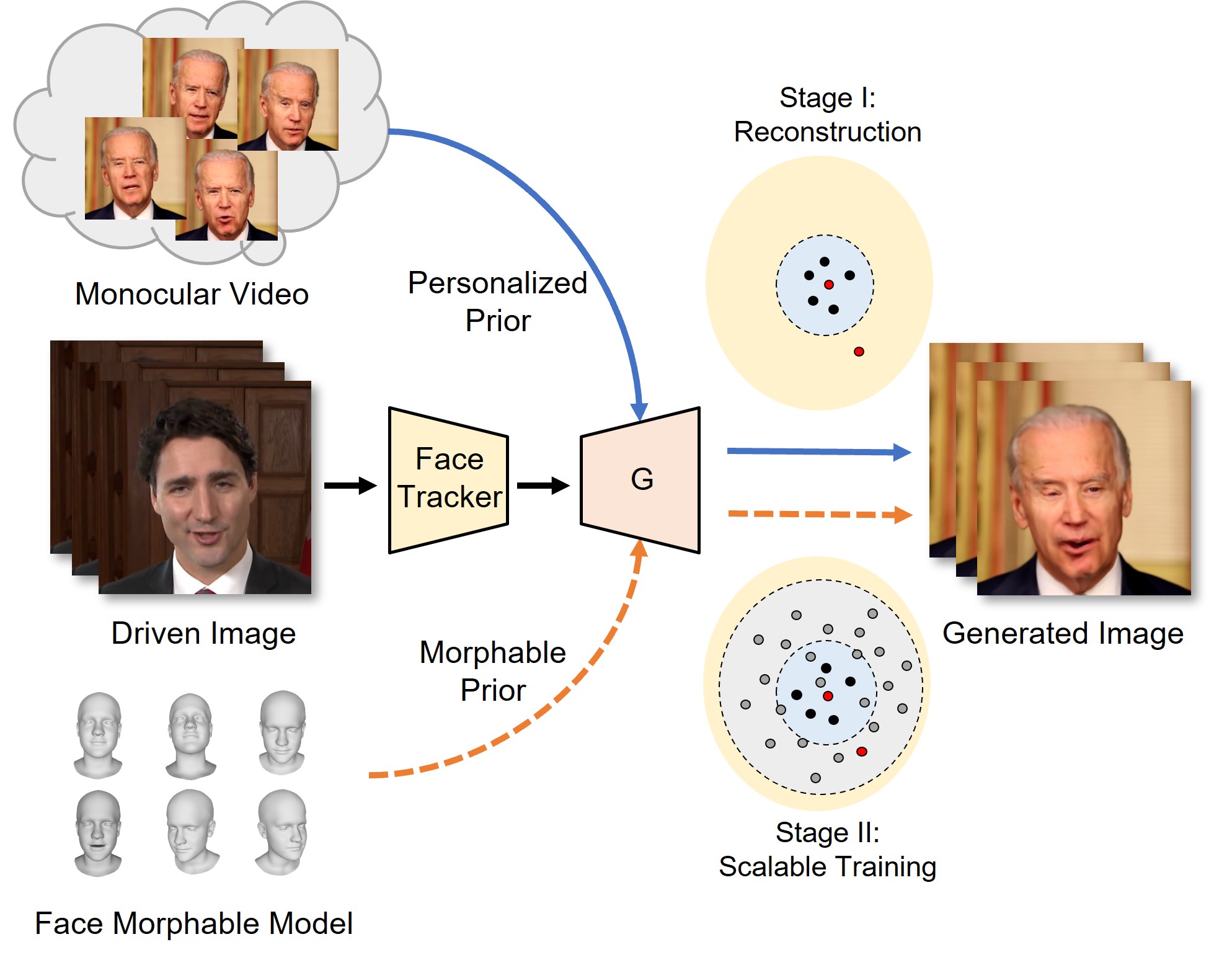}
    \caption{The proposed framework for neural portrait generation. In contrast to previous methods, we explicitly define personalized details in a monocular video as personalized prior. We also introduce a morphable prior in addition to the personalized prior for the generation of high-quality talking faces.}
    \label{img:morphable-concept}
\end{figure}

In this work, we study the problem of realistic talking face generation given only a monocular video of an actor and an input control signal. We develop our proposed method upon the existing 3D-based methods \cite{kim2018deep,gafni2021dynamic,wang2023styleavatar,doukas2023dynamic}, which achieve significant progress in terms of realism and controllability. The key insight is that a monocular video contains abundant information for face reenactment for a person, such as fine textures in various poses, expressions, and phonemes. We explicitly define the facial information in a monocular video as \textbf{personalized prior}. Despite the impressive progress of the existing methods, a person's monocular video usually contains limited 3D semantic parameters (\eg pose and expression) such that the quality of generated faces is not guaranteed for novel parameters. 

We further introduce a \textbf{morphable prior}, \textit{i.e.} facial shape prior in the format of 3DMM to tackle the generalization problem to novel parameters. Previous works \cite{blanz1999morphable,booth20163d,paysan20093d,li2017learning} demonstrate that the 3DMM can capture a large percentage of shape variations of 3D facial shapes with only a few linear bases (principal components). In this way, expressions can be decoupled with identities and poses. The morphable prior provides diverse 3D semantic parameters (\eg pose and expression) that mimic the parameters of faces in real scenes and is thus helpful for the generation of novel parameters. This also broadens the expressiveness of the trained models and enhances the quality of the generated videos. 

The aforementioned priors motivate us to propose \textbf{MyPortrait}, a novel prior-guided framework for neural portrait generation, as shown in \cref{img:morphable-concept}. In addition to the personalized prior which is obtained from a monocular video of a single person, we get the morphable prior from auxiliary data in real-life videos. We consider that these auxiliary data contain sufficient 3D semantic parameters for training. Our proposed framework supports both video-driven and audio-driven applications without altering the trunk network structure. Distinguished by whether to send the test data in training, we include an \textbf{online} and an \textbf{offline} version of our method. The online method can generate images in real-time, while the offline method can greatly improve the generation quality over the existing works. We conduct comprehensive experiments on various datasets to demonstrate the superiority of our proposed framework. In summary, our main contributions are as follows:

\begin{itemize}
    \item We design a simple, general, and flexible framework for neural portrait generation, which supports both video-driven and audio-driven facial animation given a monocular video of a person.
    \item We propose a novel prior-guided training strategy for personalized portrait generation with realistic details, which firstly combines personalized prior from a monocular video and morphable prior from face morphable space.
    \item Comprehensive experiments demonstrate that our method improves the quality of generation over the state-of-the-art methods.
    % \item We utilize two vital priors for neural portrait generation, including personalized prior and morphable prior. The former provides a rich facial prior, and the latter guides the network to repair facial details in the generated image.
    % \item We propose a prior-guided training strategy for personalized portrait generation. The proposed strategy combines personalized prior from a monocular video and morphable prior from face morphable space to improve the generation quality of the model.
    % \item We design a simple, general, and flexible network that outperforms state-of-the-art methods in comprehensive experiments even without additional manual annotations.
\end{itemize}

\section{Related work}
\label{sec:related}

% In this section, we first present related work on neural portrait generation and then introduce research on portrait generation based on generative models and personalization.

%-------------------------------------------------------------------------
\subsection{Neural Portrait Generation}

\noindent \textbf{2D-based methods.} Several 2D-based methods \cite{wiles2018x2face, siarohin2019animating, siarohin2021motion, zhao2022thin,oorloff2023robust} achieve facial pose and expression transfer by performing warping at the image or feature level. For instance, FOMM \cite{siarohin2019first} is an early method that employs motion field-based image warping to achieve motion transfer without requiring any annotation or prior information about the specific object to be animated. DaGAN \cite{hong2022depth} recovers dense 3D geometry for talking head video generation. LIA \cite{wang2022latent} employs self-supervised autoencoders to animate images by navigating in the latent space. Although only one image for the input is required in these methods, they are difficult to recover personalized facial details due to the lack of complete 3D face information in a single image.

\noindent \textbf{3D-based methods.} Several 3D structural-based methods \cite{kim2018deep, ren2021pirenderer, grassal2022neural, xu2023latentavatar,zheng2022avatar} use 3D morphable model \cite{blanz1999morphable,booth20163d,paysan20093d,li2017learning} for portrait generation. Among them, the monocular video-based portrait generation is a mainstream approach. DVP \cite{kim2018deep} proposes a generative neural network with a novel spatio-temporal architecture that enables realistic re-animation of portrait videos using a single input video. NerFace \cite{gafni2021dynamic} introduces dynamic neural radiation fields for modeling the appearance and dynamics of faces. StyleAvatar \cite{wang2023styleavatar} proposes a real-time realistic portrait avatar reconstruction method using the StyleGAN~\cite{karras2019style}. Unlike the complex processing steps of the above methods, DNP \cite{doukas2023dynamic} proposes a simple 2D coordinate-based multi-layer perception (MLP) architecture that learns the mapping from face parameter space to images, which generates controllable dynamic neural portraits. Although the above methods introduce 3DMM, the parameter space is limited in a monocular video. It leads to a degradation of the generation quality when encountering unseen face parameters. To alleviate the difficulty, we introduce a general morphable prior to enrich the face parameters space for training. The quality of the resulting videos is enhanced notably.

\begin{figure*}[t]
    \centering
    \includegraphics[width = 0.98\textwidth]{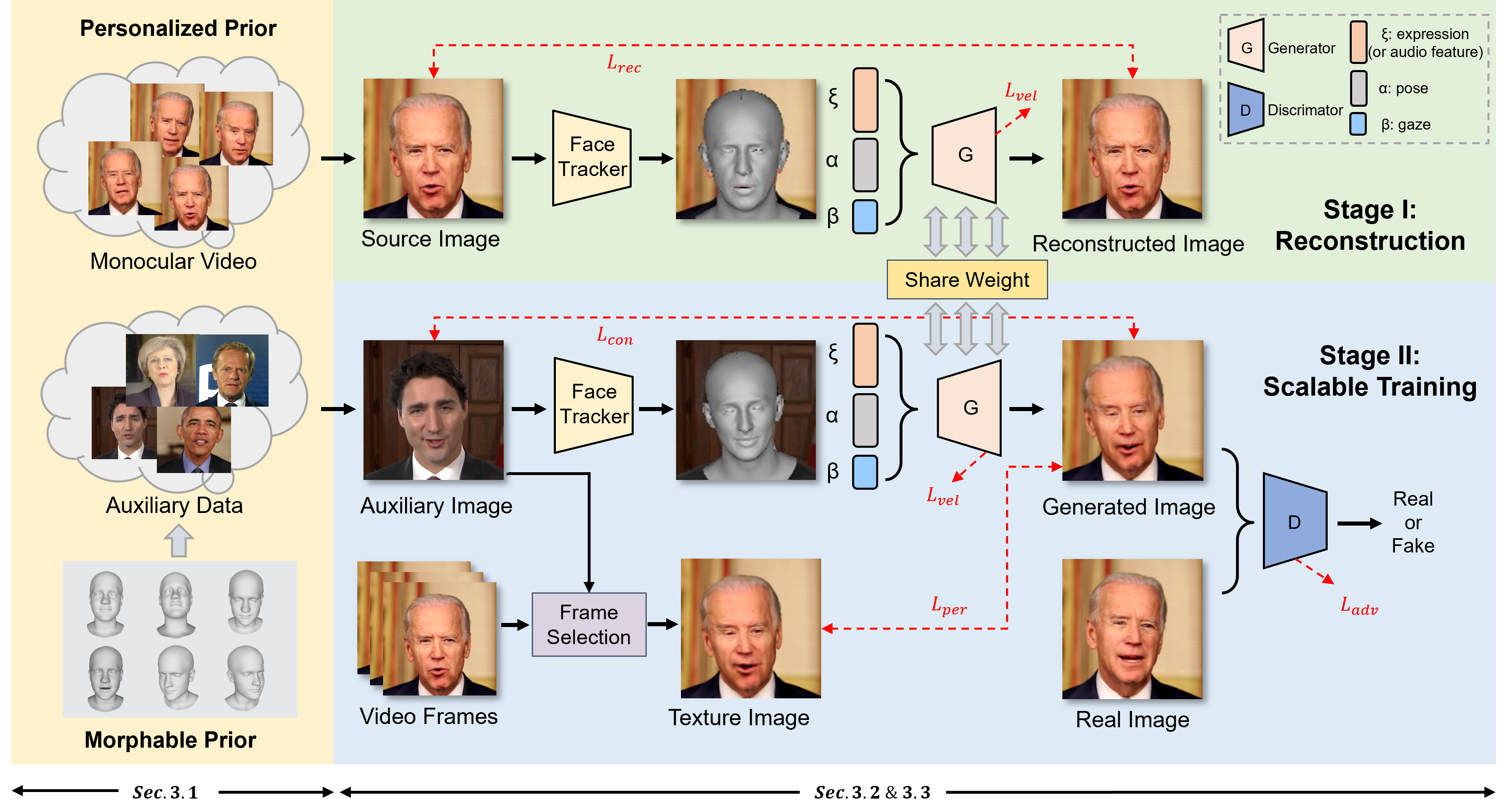}
    \caption{The overview of our method. The network first learns to reconstruct facial images from specific face parameters for a monocular video. The learned face parameter space is then extended by a scalable training strategy, which renders the method capable of generating personalized neural portraits under novel parameters.}
    \label{img:3.1:network pipeline}
\end{figure*}

%-------------------------------------------------------------------------
\subsection{Generative Model and Personalization}

Several methods \cite{yin2022styleheat, sun2023next3d, zhang2023metaportrait,ma2023otavatar,bai2023high} leverage universal generative models to repair missing facial details for reenactment. For instance, StyleHEAT \cite{yin2022styleheat} utilizes a pre-trained StyleGAN \cite{karras2019style,karras2020analyzing,karras2021alias} model to enable high-resolution controllable talking face generation. Next3D \cite{sun2023next3d} employes a pre-trained EG3D \cite{chan2022efficient} model to create a 3D-aware head avatar. These methods obtain fine-grained control over face motion but suffer from missing personalized facial details. To address this limitation, Mystyle \cite{nitzan2022mystyle} uses a small number of portrait images of a specific person to fine-tune the weights of the pre-trained StyleGAN face generator, thereby forming a local, low-dimensional, and personalized space in the latent space. Mystyle++ \cite{zeng2023mystyle++} further proposes an approach to obtain a personalized generative prior with explicit control over a set of attributes. These methods utilize the latent space of a pre-trained generative model for better fine-grained face details. In this work, our method introduces a similar strategy for fine-tuning the results at test time. In a departure from the existing works, we extend the face parameter space by a morphable prior in training for better generalization performance. The architecture of our method is also different from the Mystyle \cite{nitzan2022mystyle} and Mystyle++ \cite{zeng2023mystyle++}.

\section{Method}
\label{sec:method}

 In this section, we first introduce two priors for portrait generation, including the personalized prior and the morphable prior. Then, our efficient network structure and loss functions are presented. Finally, we elaborate the training strategy for personalized portrait generation. An overview of our method is presented in \cref{img:3.1:network pipeline}.

% Outline
% 3DMM中的变形先验
% 先验引导的框架
% 高效的网络结构

%-------------------------------------------------------------------------
\subsection{Priors for portrait generation}

\textbf{Personalized Prior.} Some existing methods \cite{kim2018deep,gafni2021dynamic,doukas2023dynamic,wang2023styleavatar} use the rich prior information of faces, such as finely textured details under different poses, expressions, and phonemes in a monocular video to generate realistic talking faces. Due to the powerful learning capacity of deep neural networks, these methods are able to generate fake faces that are indistinguishable from real ones. The key to success is that a monocular video contains all the detailed texture information, both globally and locally for a person under various controllable parameters, \textit{e.g.} expressions in video-driven facial animations and phonemes in audio-driven facial animations. The texture information can be stored in the weights of a well-trained network to repair the missing facial details in certain semantics. In this work, we explicitly define the detailed texture information in a monocular video as \emph{personalized prior}.  

\noindent \textbf{Morphable Prior.} Our goal is to generate realistic facial videos of a person under various semantics which are related to varying shapes and poses of a face. The deformation of shapes is commonly caused by expressions and phonemes, depending on video-driven and audio-driven applications, respectively. Taking the video-driven applications as an example\footnote{We can replace the expressions by the phonemes in audio-driven applications.}, we define the morphable parameter space of all possible poses and expressions of a single person as $\bf{M}$. We call this the \emph{morphable prior} coupled with the high-level semantics. 

\begin{figure}[t]
    \centering
    \includegraphics[width = 0.47\textwidth]{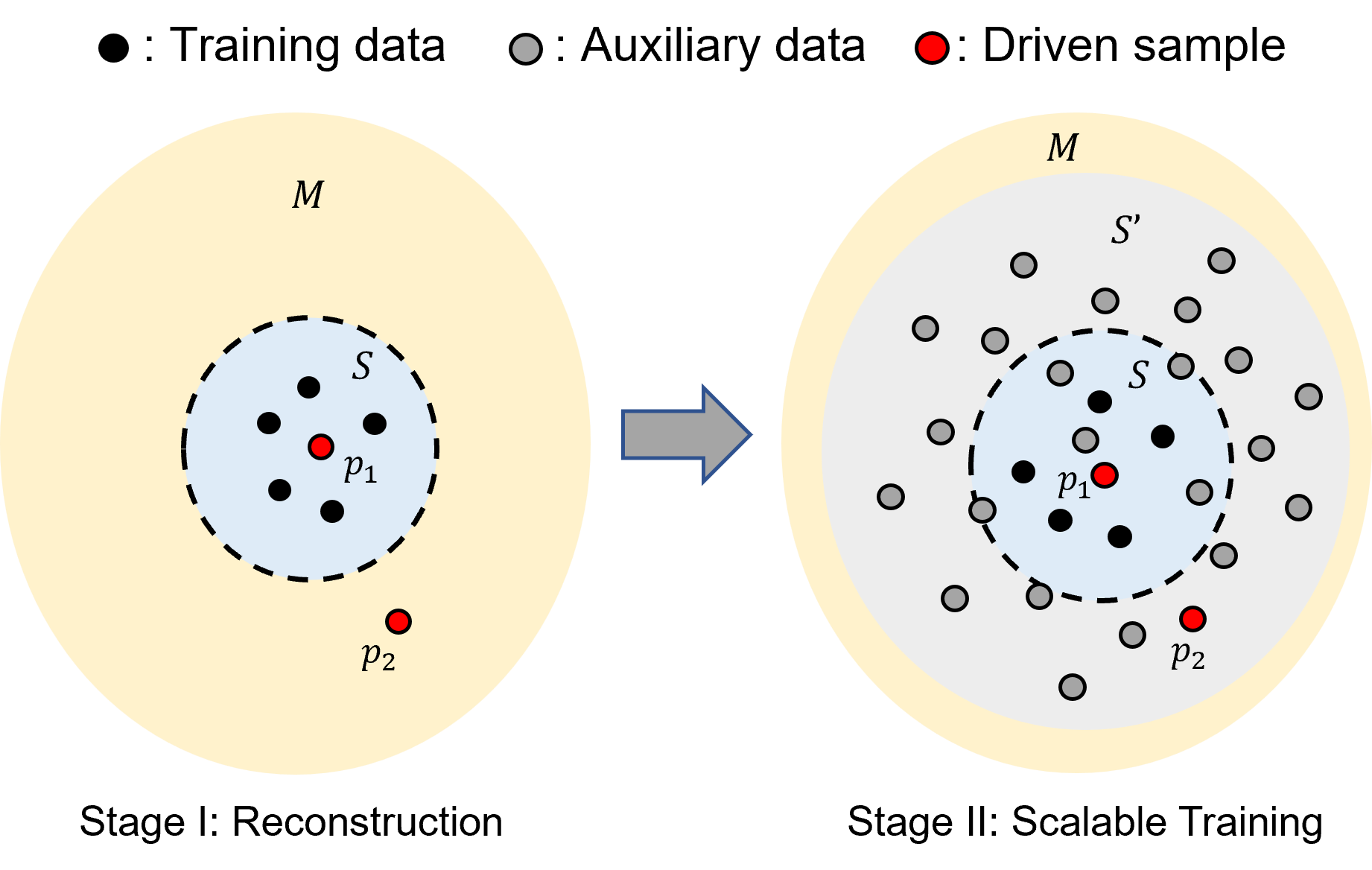}
    \caption{Variations of face parameter space for two training stages. In the first stage, the network learns personalized prior from a monocular video. In the second stage, we introduce morphable prior to extend the face parameter space. $p_1$ and $p_2$ show two examples which are inside and outside the original parameter space $\bf{S}$ of the input monocular video, respectively. After the second stage, $p_2$ is covered by the extended parameter space $\bf{S'}$.}
    \label{img:morphable_space-concept}
\end{figure}

Ideally, we hope that a trained model for a monocular video is generalizable to all possible semantic parameters. However, this is not achievable due to the limited semantics of a monocular video in practical applications. We define the limited parameter space provided in the monocular video as $\bf{S}$, which is smaller than $\bf{M}$, as
\begin{equation}
    \centering
    \bf{S} \subset \bf{M}.
\end{equation}
Most existing models~\cite{grassal2022neural,zheng2022avatar,wang2023styleavatar} can only animate the input parameters in the subspace $\bf{S}$. The generation quality is commonly degraded for unseen input parameters in $\bf{M}/\bf{S}$. To overcome this limitation, we introduce a universal facial prior in terms of the poses and expressions of the 3DMM \cite{blanz1999morphable,booth20163d,paysan20093d,li2017learning}. 
The prior captures general domain knowledge about facial shape variations, which allows any face to be represented as a linear combination of a finite basis set of face prototypes. We then sample the morphable parameter space in $\bf{M}$ to extend the input parameter space $\bf{S}$. The extended parameter space are denoted by $\bf{S'}$. Finally, the trained model can generate high-quality videos in the extended parameter space $\bf{S'}$, which approaches the ideal morphable parameter space $\bf{M}$, as
\begin{equation}
    \centering
    \bf{S} \subset \bf{S'} \to \bf{M}.
\end{equation}
The process is illustrated in \cref{img:morphable_space-concept}. 

In our implementation, we sample the morphable parameters from real-life videos with a state-of-the-art 3D reconstruction method~\cite{feng2021learning}. We call these real-life videos as \emph{auxiliary data}, which provides not only various morphable parameters but also temporal sequences for training. After all, the frames from auxiliary data contain smoothly varying parameters and contribute to temporally coherent results. In addition, the extended parameter space $\bf{S'}$ approaches the ideal morphable space $\bf{M}$ as the auxiliary data increase.

%-------------------------------------------------------------------------
\subsection{Network structure and loss functions}

\begin{figure}[t]
    \centering
    \includegraphics[width = 0.47\textwidth]{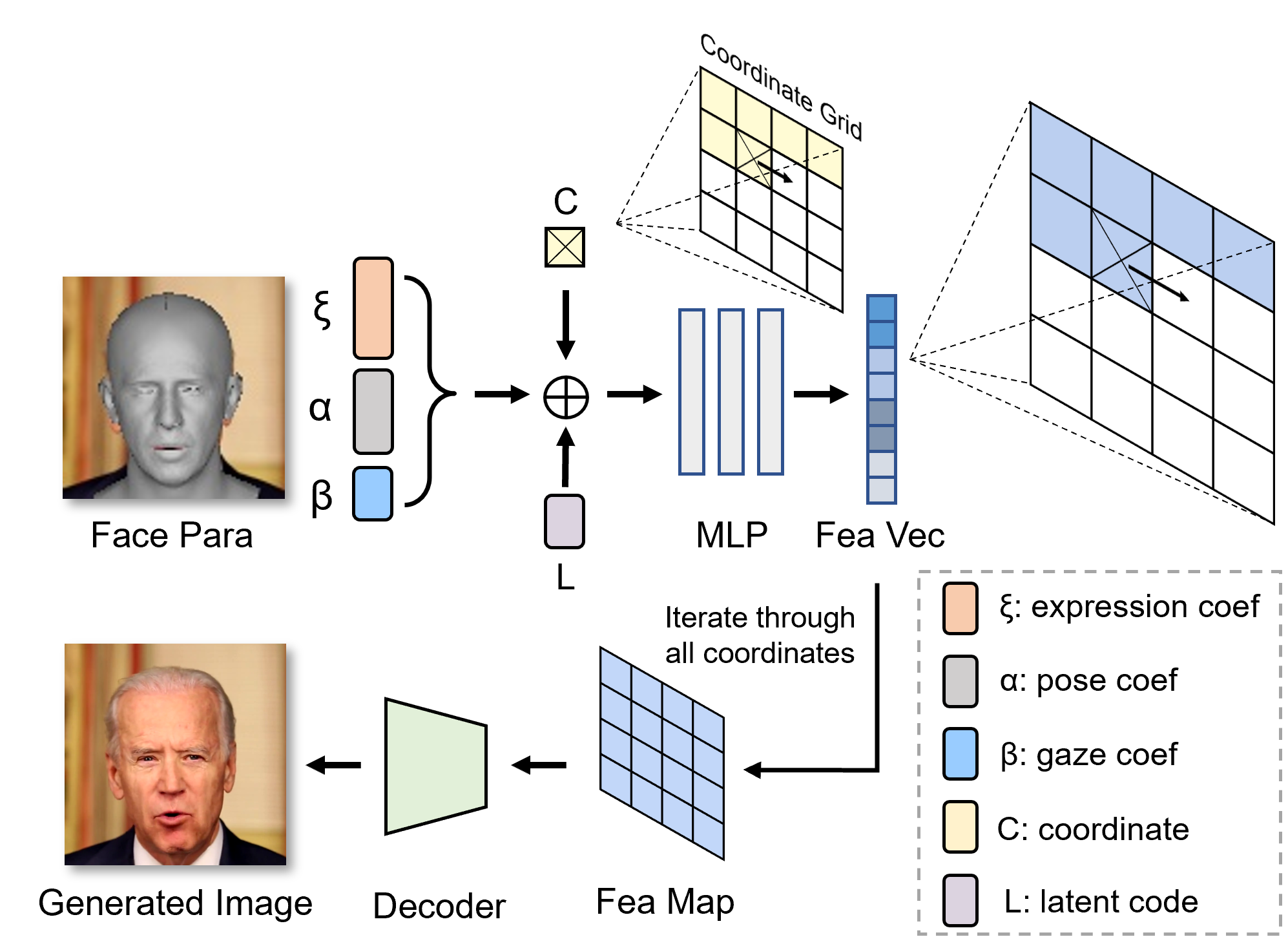}
    \caption{The illustration of the generator network. The generator consists of a 2D coordinate-based MLP and a CNN-based decoder.}
    \label{img:Generator-network}
\end{figure}

We use a simple 2D coordinate-concatenated MLP and CNN-based decoder network architecture of which the main branch is similar to the DNP \cite{doukas2023dynamic}. This considers the training efficiency in case of the increase of auxiliary data compared to a monocular video. Unlike DNP, we use bilinear interpolation rather than transposed convolution for upsampling because we observe that transposed convolution produces checkerboard artifacts \cite{odena2016deconvolution}. This also reduces the computational cost. We utilize per-image learnable latent code to model information in the image that is difficult to describe in terms of face parameters, including factors such as facial details, bodily motion, and illumination changes. Besides the pose and expression parameters of the 3DMM, we introduce gaze as an input parameter to control the gaze angle of the generated image in video-driven tasks. The structure of the generator is illustrated in \cref{img:Generator-network}. In addition, we also include three auxiliary networks in training to enhance the generation quality. Firstly, we adopt a patch-based GAN architecture \cite{goodfellow2014generative,isola2017image,zhu2017unpaired} to force the generated faces to be indistinguishable to truth faces. Secondly, we employ the pre-trained VGGNet \cite{simonyan2014very,johnson2016perceptual} for adding perceptual loss as a common routine in low-level feature generation methods \cite{wang2023styleavatar,bai2023high}. Finally, we also include a feature extraction network for parameter consistency between the generated face and the driven one. \footnote{The details for the network structure are included in the supplementary material.}

% Benefiting from the simple network structure and the introduction of auxiliary data, we only need to replace the expression coefficients with audio features for audio-driven talking face generation. We first process the audio signal with the pre-trained DeepSpeech \cite{hannun2014deep,amodei2016deep} and then use audio network \cite{guo2021ad} processing to get the audio features. For more details about the network structure, please refer to the supplementary material.

\noindent \textbf{Loss Function.} A variety of loss functions are carefully designed to train the network efficiently, as follows (also refer to \cref{img:3.1:network pipeline}).

(a) \textit{Reconstruction loss} ensures the generated image $G(c)$ to match the ground truth image $y$ as
\begin{equation}
    \centering
    L_{rec}=\Vert |y-G(c)| \Vert_2,
    \label{eq:recons_loss}
\end{equation}
where $G$ denotes the generator network taking the face parameter $c$ as input.

(b) \textit{Perceptual loss} ensures semantic feature consistency between the generated image and the ground truth image as
\begin{equation}
    \centering
    L_{per}= \sum_{i=1}^I \Vert N_i(t)-N_i(G(c)) \Vert_1,
    \label{eq:percep_loss}
\end{equation}
where $N_i(\cdot)$ represents the $i^{th}$ channel feature extracted from a specific VGG-19 layer, $t$ is the texture image obtained by performing frame selection based on the nearest neighbor of the face parameter, and $I$ is the number of feature channels in this layer. 

(c) \textit{Consistency loss} ensures face parameters (\eg expressions, and poses) consistency between the generated face and the driven one, as
\begin{equation}
    \centering
    L_{con}=\Vert c-\Phi(G(c)) \Vert_2,
    \label{eq:cons_loss}
\end{equation}
where we implement $\Phi(\cdot)$ with the DECA \cite{feng2021learning} model.

(d) \textit{Adversarial loss} consists of two components: $L_{adv}^D$ for the discriminator and $L_{adv}^G$ for the generator to force the distribution of generated images to align with the real ones as:
\begin{equation}
        L_{adv}^D = log(D(G(c)))+log(1-D(y)),
        \label{eq:adv_d_loss}
\end{equation}
and
\begin{equation}
        L_{adv}^G = log(1-D(G(c))).
        \label{eq:adv_g_loss}
\end{equation}

(e) \textit{Velocity loss} ensures coherence between adjacent frames as 
\begin{equation}
    \centering
    L_{vel}=(\Vert v_{i}\Vert_2+\Vert v_{i+1}\Vert_2)+\Vert v_{i}-v_{i+1}\Vert_2,
    \label{eq:velocity_loss}
\end{equation}
where $i$ represents the $i^{th}$ frame, and $v_i$ and $v_{i+1}$ represent the corresponding latent codes of the adjacent frames. The two terms in \cref{eq:velocity_loss} encourage sparsity and temporal consistency of the latent code, respectively.

Finally, the full objective functions are 
\begin{equation}
    \centering
    L_D = L_{adv}^D,
    \label{eq:d_loss}
\end{equation}
and
\begin{equation}
    \centering
     L_G = L_{adv}^G+\alpha_{1} L_{rec}+\alpha_{2} L_{per} + \alpha_{3} L_{con}+\alpha_{4} L_{vel},
     \label{eq:g_loss}
\end{equation}
for the discriminator and the generator, respectively. Here, $\alpha_{1}$, $\alpha_{2}$, $\alpha_{3}$, and $\alpha_{4}$ are weights for different terms.

%-------------------------------------------------------------------------
\subsection{Prior-guided training strategy}

In this section, we propose a prior-guided \textit{two-stage} training strategy to effectively make use of the aforementioned two priors for personalized portrait generation with high-quality details. 

\noindent \textbf{Stage I: Reconstruction Training.} In the first stage, we perform reconstruction training on a monocular video. Specifically, we use a face tracker to extract face parameters in the monocular video. We employ DECA \cite{feng2021learning} and DeepSpeech \cite{hannun2014deep,amodei2016deep} for video-driven and audio-driven tasks as the face tracker, respectively. The output parameters (expressions, poses, or phonemes) of the face tracker are used as the input parameters of our network. During this training stage, we only use the reconstruction loss in \cref{eq:recons_loss} and the velocity loss in \cref{eq:velocity_loss}.

\begin{figure}[t]
    \centering
    \includegraphics[width = 0.47\textwidth]{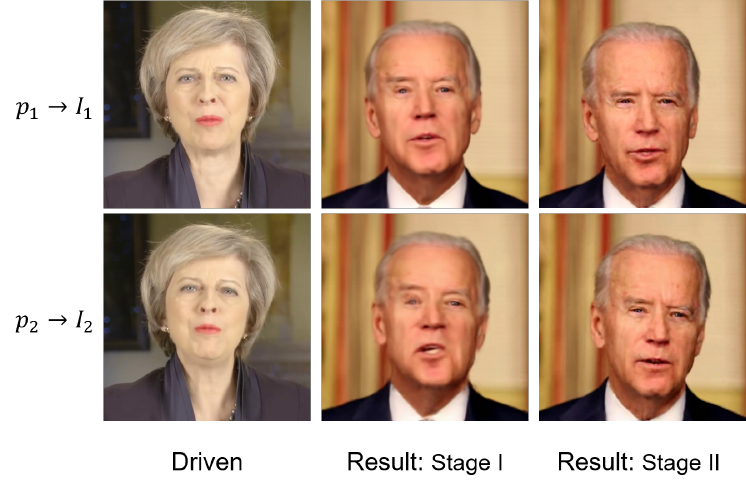}
    \caption{Two examples (also refer to \cref{img:morphable_space-concept}) for $p_1$ and $p_2$ for neural portrait generation in the two stages. The top examples show good reconstruction in the first stage. The bottom examples show some artifacts in the first stage, such as the inaccurate mouth and blurred eyes, while they are well-fixed by the second stage.}
    \label{img:morphable_res-concept}
\end{figure}

Now our goal is to learn personalized details from the monocular video, in which we explicitly define the information as the personalized prior. During training, the mapping of different face parameters to reconstructed images is learned. Benefiting from a fixed identity and background, the monocular video can recover the personalized details of the given identity. Meanwhile, the irrelevant content, such as the identity and background, remains unchanged. \cref{img:morphable_res-concept} shows two examples inside and outside the parameter space of the monocular video, respectively (also refer to $p_1$ and $p_2$ in \cref{img:morphable_space-concept}). The limitation is that the generation quality is not guaranteed for novel parameters that are located outside the parameter space $S$ of the monocular video. 

\noindent \textbf{Stage II: Scalable Training.} In the second stage, we aim to incorporate the morphable prior into training for generating realistic faces with novel parameters. We extract face parameters from auxiliary data and feed them into training. 

% We use the frames of the monocular video as performance images, and then we perform face matching based on the similarity of the face parameters between these images and the auxiliary image. Finally, the most similar one is used as the texture image.

Our purpose now is to generate high-quality faces even if the parameter is not included in the monocular video. Since the novel parameters are provided by auxiliary data, the corresponding ground truth images for the identity in the monocular video are missing. In this stage, the loss terms in \cref{eq:percep_loss,eq:cons_loss,eq:adv_d_loss,eq:adv_g_loss}, together with the auxiliary networks are employed for the novel parameters extracted from auxiliary data. As a result, the trained network is capable of fixing the missing details conditioned on the novel parameters, thereby improving the quality of generated faces notably as in \cref{img:morphable_res-concept} (the bottom example).

\cref{img:morphable_space-concept} and \cref{img:morphable_res-concept} also show that the parameter space is broadened after scalable training, as our desired effect by the introduction of the proposed morphable prior. In practice, we can extract the semantic parameters of the driven video and feed them into scalable training. We find that the quality of the generated faces is greatly enhanced in this way. This is in fact a similar strategy for fine-tuning the results at test time \cite{gandelsman2022test,sun2023learning} at the cost of sacrificing real-time performance. We distinguish our proposed method by an \textbf{online} version and an \textbf{offline} version based on whether to send the test data in training. The online method can generate images in real-time, while the offline method can greatly improve the generation quality over existing works.

\section{Experiments}
\label{sec:exp}

%-------------------------------------------------------------------------
\subsection{Experiment Setup}

\begin{table}[t]
    \centering
    \resizebox{0.45\textwidth}{!}{
    \begin{tabular}{lcccccc} 
    \hline
    Method & Portrait & L1($\downarrow$) & LPIPS($\downarrow$) & FID($\downarrow$)   \\ 
    \hline
                 & ID.1     & 0.052 & 0.078    & 54.184   \\
    {FOMM \cite{siarohin2019first}}       & ID.2     & 0.063 & 0.134    & 137.60  \\
                 & ID.3     & 0.039 & 0.090    & 79.780   \\ 
    \hline
                 & ID.1     & \textbf{0.037} & \textbf{0.056} & 63.514   \\
    {NerFACE \cite{gafni2021dynamic}}    & ID.2     & 0.048 & 0.118    & 102.26  \\
                 & ID.3     & 0.023 & \textbf{0.025}    & 41.268   \\ 
    \hline
                 & ID.1     & 0.046 & 0.112    & 23.975   \\
    {DNP \cite{doukas2023dynamic}}        & ID.2     & 0.032 & 0.081    & 20.834   \\
                 & ID.3     & 0.020 & 0.040    & 21.536   \\
    \hline
                 & ID.1     & 0.045 & 0.102    & \textbf{15.441}   \\
    {Ours}       & ID.2     & \textbf{0.031} & \textbf{0.073}    & \textbf{16.920}  \\
                 & ID.3     & \textbf{0.020} & 0.037    & \textbf{17.894}   \\ 
    \hline
    \end{tabular}
    }
    \caption{Comparison to self-reenactment methods on the NerFace dataset. The bold results are the best.}
    \label{tab:Comp-reconstruction}
\end{table}

\noindent \textbf{Dataset.} Our approach trains neural portraits on monocular videos, creating a new model for each identity-specific video. The experimental data come from Head2Head \cite{koujan2020head2head,doukas2021head2head++} and NerFACE \cite{gafni2021dynamic}. We extract and store the expression, pose, phoneme, and gaze parameters of the facial videos for data preprocessing. We also pre-crop the face images for the actor in the monocular video, keeping a fixed background and camera viewpoint. Please refer to the supplementary material for more details.

\noindent \textbf{Metric.} Referring to existing monocular video-based methods \cite{kim2018deep,gafni2021dynamic,wang2023styleavatar,doukas2023dynamic}, we evaluate the performance of our method in self-reenactment and cross-reenactment experiments. In the self-reenactment experiment, we use L1-distance (L1) and Learned Perceptual Image Patch Similarity (LPIPS) \cite{zhang2018unreasonable} to quantitatively assess the portrait reconstruction with respect to ground truths. Additionally, we use the
Frechet Inception Distance (FID) \cite{heusel2017gans}, a metric for the realism of generated images to evaluate the difference between the feature distributions of generated and real images. In the cross-reenactment experiment, Cosine Similarity (CSIM) of identity embedding is used to assess the preservation of the target's identity information in the generated images, with the assistance of ArcFace \cite{deng2019arcface}. Furthermore, we refer to PIRender \cite{ren2021pirenderer} to assess the semantic consistency of the driven signal using the Average Expression Distance (AED) and the Average Pose Distance (APD).

\subsection{Comparison to the State of the Art}
We compare the performance of our method with state-of-the-art methods, including FOMM \cite{siarohin2019first}, StyleHEAT \cite{yin2022styleheat}, NerFACE \cite{gafni2021dynamic}, and DNP \cite{doukas2023dynamic}. FOMM and StyleHEAT are 2D-based methods for face reenactment based on a single reference image. NerFACE and DNP are 3D-based methods that leverage a monocular video to generate neural portraits.

\begin{figure}[t]
    \centering
    \includegraphics[width = 0.47\textwidth]{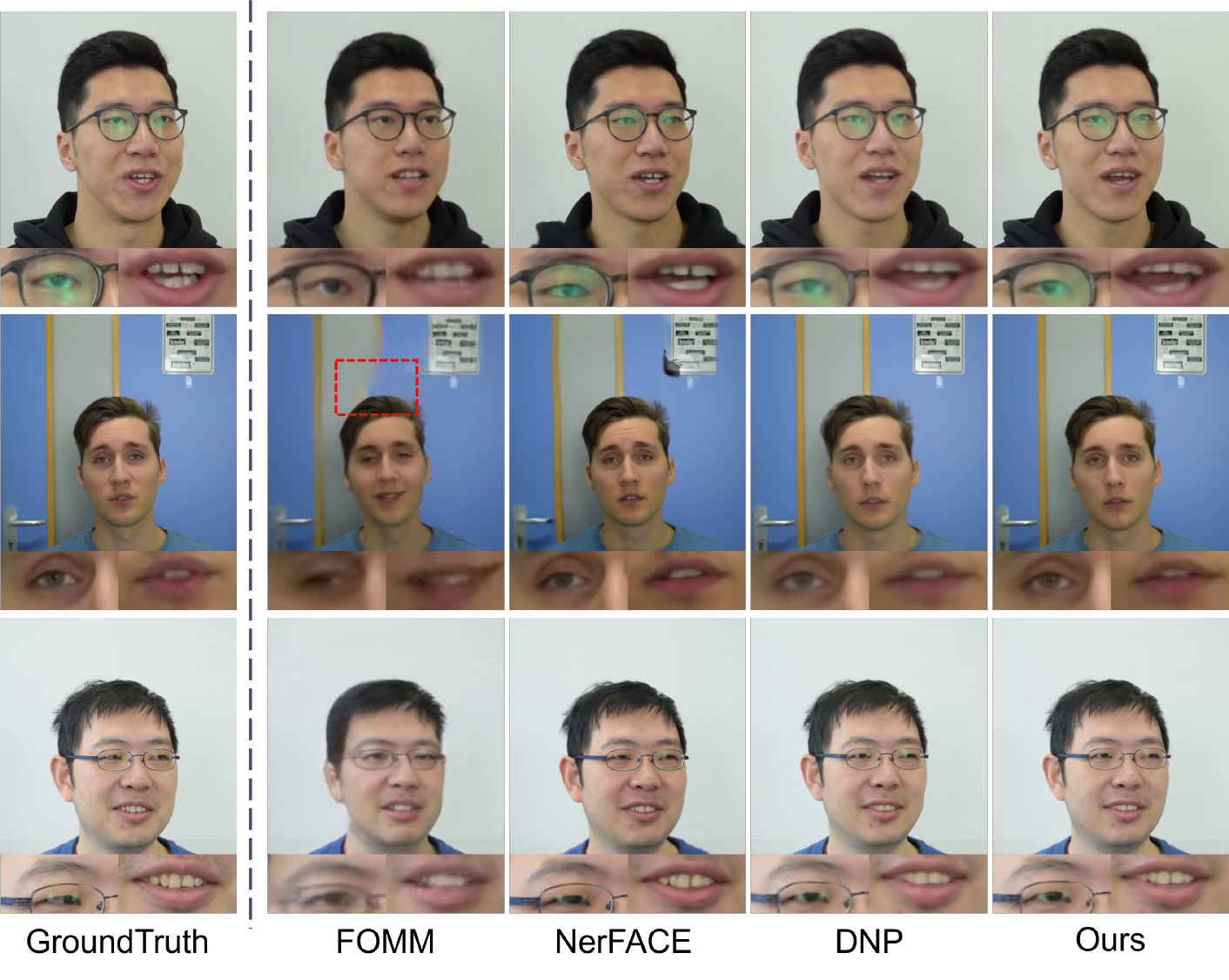}
    \caption{Visual comparison to self-reenactment methods. Please zoom in to see the details.}
    \label{img:self-reenactment}
\end{figure}

\begin{table*}
\centering
\resizebox{0.99\textwidth}{!}{
    \begin{tabular}{l|cccc|cccc|cccc} 
    \hline
     \multirow{2}{*}{Method}     & \multicolumn{4}{c|}{Target: Biden~ Driven: May} & \multicolumn{4}{c|}{Target: Trudeau~ Driven: tusk} & \multicolumn{4}{c}{Target: May~ Driven: Trudeau}  \\ 
    \cline{2-13}
         & CSIM(↑) & APD(↓) & AED(↓) & FID(↓)                & CSIM(↑) & APD(↓) & AED(↓) & FID(↓)                   & CSIM(↑) & APD(↓) & AED(↓) & FID(↓)                  \\ 
    \hline
    % FOMM \cite{siarohin2019first}            & 0.475   & 0.018  & 0.255  & 21.086                & 0.574   & \textbf{0.015}  & 0.285  & 10.444                   & 0.681   & \textbf{0.028}  & 0.220  & 37.244                  \\
    StyleHEAT \cite{yin2022styleheat}      & 0.358   & 0.014  & \textbf{0.184}  & 61.285                & 0.381   & \textbf{0.015}  & \textbf{0.253}  & 93.360                   & 0.158   & \textbf{0.028}  & 0.209  & 132.704                 \\
    DNP \cite{doukas2023dynamic}            & 0.278   & \textbf{0.011}  & 0.200  & 23.964                & 0.385   & 0.015  & 0.288  & 116.512                  & 0.508   & 0.030  & 0.252  & 96.323                  \\
    % \hline
    Ours (online)  & 0.327   & 0.015  & 0.204  & 22.500                & 0.573   & 0.028  & 0.416  & 30.429                   & 0.664   & 0.039  & 0.275  & 68.745                  \\
    Ours (offline) & \textbf{0.514}   & 0.012  & 0.214  & \textbf{8.553}                 & \textbf{0.634}   & 0.018  & 0.291  & \textbf{5.513}                    & \textbf{0.749}   & 0.033  & \textbf{0.203}  & \textbf{15.590}                  \\
    \hline
    \end{tabular}
    }
    \caption{Comparison to cross-reenactment methods on the Head2Head dataset. The bold results are the best.}
    \label{tab:Comp-reenactment}
\end{table*}

\noindent \textbf{Self-Reenactment.} We conduct self-reenactment experiments on the NerFace dataset to compare with the existing methods. The quantitative experimental results are presented in \cref{tab:Comp-reconstruction}. Compared to state-of-the-art methods, our method achieves superior performance in all metrics. The reconstruction loss allows us to achieve a lower L1 distance. Additionally, by optimizing the perceptual loss and the adversarial loss, the feature distribution of the images generated by our method aligns better with that of the real images, resulting in a lower LPIPS and FID. Visual comparison with self-reenactment methods is illustrated in \cref{img:self-reenactment}. FOMM \cite{siarohin2019first} uses only one reference image for motion transfer, leading to the lack of facial prior in the generated images. The warping strategy based on the motion field also causes distortions in the background, as shown by the red box in \cref{img:self-reenactment}, middle row. NerFACE \cite{gafni2021dynamic} assumes that the head pose parameters are consistent with the camera viewpoints, resulting in inconsistencies in the generated torsos. DNP \cite{doukas2023dynamic} achieves good performance in terms of generation quality after simplifying 3D to 2D scenes. However, it is limited by the different distributions of training and testing data, which results in missing facial details in the generated images. In contrast, our method introduces morphable prior to extend the face parameter space in training, thus greatly improving the generation quality.

\begin{figure}[t]
    \centering
    \includegraphics[width = 0.47\textwidth]{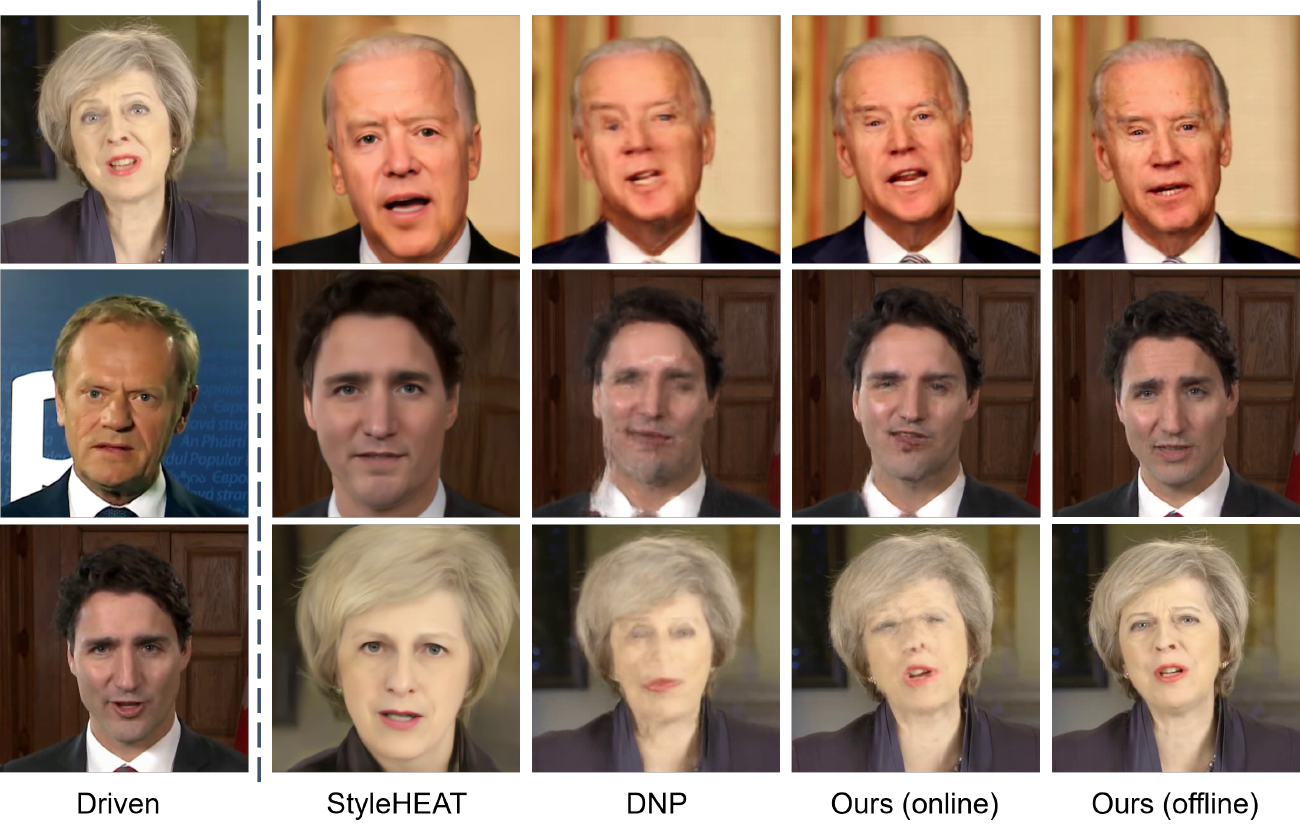}
    \caption{Visual comparison to cross-reenactment methods. Please zoom in to see the details.}
    \label{img:cross-reenactment}
\end{figure}

% \begin{table}[t]
%     \centering
%     \resizebox{0.47\textwidth}{!}{
%     \begin{tabular}{cccccccc} 
%     \hline
%     Method & CSIM($\uparrow$) & APD($\downarrow$) & AED($\downarrow$) & FID($\downarrow$)   \\ 
%     \hline
%     \multirow{2}{*}{FOMM}   & 0.587 & 0.048    & 0.176  & 54.840  \\
%                             & 0.472 & \textbf{0.019}    & 0.266  & 19.689  \\
%     \hline
%     \multirow{2}{*}{StyleHEAT} & 0.432 & \textbf{0.015}    & 0.187  & 171.64  \\
%                             & 0.380 & 0.036    & 0.216  & 57.835  \\
%     \hline
%     \multirow{2}{*}{DNP}       & 0.603 & 0.017    & 0.143  & 33.178   \\
%                             & 0.435 & 0.030    & 0.246  & 16.711  \\
%     \hline
%     \multirow{2}{*}{Ours}      & \textbf{0.603} & 0.018    & \textbf{0.137}  & \textbf{23.220}  \\
%                                 & \textbf{0.473} & 0.025    & \textbf{0.207}  & \textbf{11.119}  \\
%     \hline
%     \end{tabular}
%     }
%     \caption{Comparison to cross-reenactment methods on the NerFace and Head2Head dataset. The bold results are the best.}
%     \label{tab:Comp-reenactment}
% \end{table}

\noindent \textbf{Cross-Reenactment.} We also conduct cross-reenactment experiments on the NerFace dataset to compare with the existing methods. It should be noted that cross-reenactment is more difficult than self-reenactment for monocular video-based portrait generation methods. The training data and testing data of cross-reenactment come from different identities, which causes significant differences in the face parameter distributions. \cref{tab:Comp-reenactment} shows the results of the quantitative experiments on cross-reenactment. Our offline method achieves the best CSIM and FID under various experimental settings. It should be noted that our method aligns the generated images to the real ones at the expense of certain face parameter consistency, which causes our method to be slightly worse than StyleHEAT \cite{yin2022styleheat} in terms of APD and AED, but with higher generation quality. The visual comparison with the cross-reenactment methods is presented in \cref{img:cross-reenactment}. Although StyleHEAT \cite{yin2022styleheat} uses universal face prior from other datasets to recover facial details, the lack of personalized prior makes it difficult to generate personalized facial images. Affected by the inconsistent distribution of face parameters between the training and driving data, unexpected artifacts appear in the generated images with DNP \cite{doukas2023dynamic}. Our online method mitigates this situation after introducing morphable prior. Further, after including the driven data in training, our offline method generates portraits with significantly improved quality.

\begin{table}[t]
    \centering
    \resizebox{0.4\textwidth}{!}{
    \begin{tabular}{lccccc} 
    \hline
    Method & L1($\downarrow$) & LPIPS($\downarrow$) & FID($\downarrow$)    \\ 
    \hline
    {SadTalker \cite{zhang2023sadtalker}} & 0.106 & 0.365 & 86.517  \\
    {DNP \cite{doukas2023dynamic}}      & 0.035 & 0.074    & 17.601   \\
    {Ours}     & \textbf{0.035} & \textbf{0.065} & \textbf{10.880}  \\ 
    \hline
    \end{tabular}
    }
    \caption{Comparison to audio-driven reenactment methods. The bold results are the best.}
    \label{tab:audio-driven}
\end{table}

\begin{figure}[t]
    \centering
    \includegraphics[width = 0.45\textwidth]{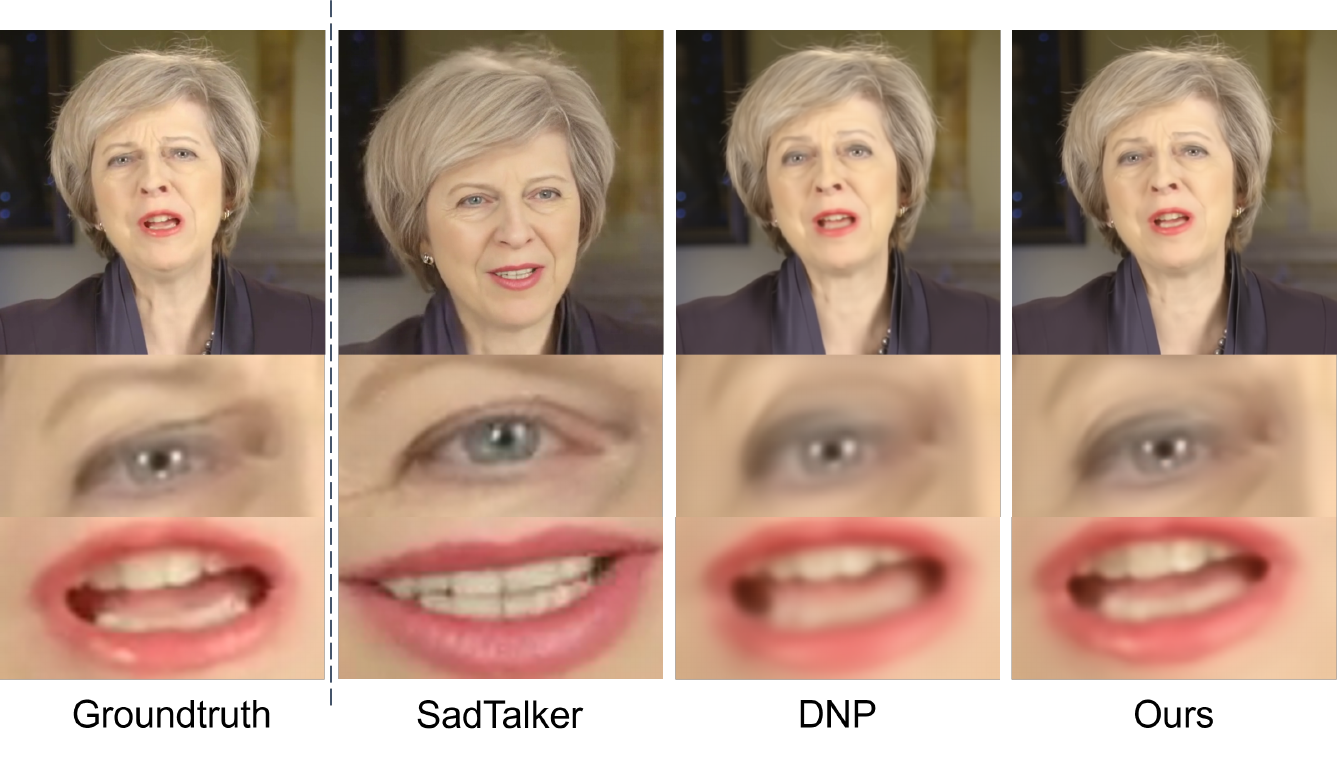}
    \caption{Visual comparison to audio-driven methods. Please zoom in to see the details.}
    \label{img:comp-audio-driven}
\end{figure}

\noindent \textbf{Audio-driven Reenactment.} Existing audio-driven face reenactment methods \cite{zhang2023sadtalker,xu2022designing,yang2022face2face,xing2023codetalker} have made significant progress. Although our framework makes improvements primarily for the video-driven method, it also achieves certain performance gains for the audio-driven applications. We include SadTalker \cite{zhang2023sadtalker} and DNP \cite{doukas2023dynamic} for comparison. SadTalker is an advanced method for generating a realistic video of a talking face based on an input piece of audio and a face image. DNP can also be used for audio-driven tasks. Quantitative results are shown in \cref{tab:audio-driven}. SadTalker, despite its ability to generate a better lip-synchronized video with the help of facial information learned from a large amount of data, is limited by the lack of information about the target face, resulting in the inability to portray personalized details. Overall, our method achieves the best generation quality. \cref{img:comp-audio-driven} presents qualitative results, suggesting that the morphable prior introduced via 3DMM is valid for improving the result of audio-driven reenactment.

% \noindent \textbf{Execution Speed.} Benefiting from the simple network structure, our method enables real-time inference. The evaluation results of the execution speed are presented in \cref{tab:speed-test}. Our method generates an image with $512 \times 512$ resolution in 15 ms, achieving the best results compared to state-of-the-art methods.

% \begin{table}[t]
%     \centering
%     % \resizebox{0.45\textwidth}{!}{
%     \begin{tabular}{lcccccc} 
%     \hline
%     \multirow{2}{*}{Method} & $256 \times 256$ & $512 \times 512$   \\
%      & time (fps) &  time (fps)   \\
%     \hline
%     {AD-NeRF \cite{guo2021ad}}    & -- & 9630 (0.10)   \\
%     {NerFACE \cite{gafni2021dynamic}}    & -- & 8465 (0.12)   \\
%     {DVP \cite{kim2018deep}}       & 65 (15.4) & 196 (5.1)   \\
%     {HeadGAN \cite{doukas2021headgan}}  & 41 (24.5) & --   \\
%     {FOMM \cite{siarohin2019first}}     & 21 (47.2) & --   \\
%     {DNP \cite{doukas2023dynamic}}      & \textbf{11 (90.9)} & 31 (32.3)   \\
%     {Ours}     & 14 (71.4) & \textbf{15 (66.7)}   \\ 
%     \hline
%     \end{tabular}
%     % }
%     \caption{Comparison of the execution time between our generative model and related methods. Time is reported in milliseconds (msec). Please note that all reported numbers refer to the forward pass time of models during inference, without considering data pre-processing.}
%     \label{tab:speed-test}
% \end{table}

\begin{table}[t]
    \centering
    \resizebox{0.45\textwidth}{!}{
    \begin{tabular}{ccccc}
    \hline
    Method & CSIM($\uparrow$) & APD($\downarrow$) & AED($\downarrow$) & FID($\downarrow$)   \\ 
    \hline
    % {DNP}           & 0.403 & 0.028 & 0.245 & 23.700  \\
    % \hline
    {wo $L_{gan}$}   & 0.437 & 0.025 & 0.255 & 13.265   \\
    {wo $L_{per}$}   & 0.459 & \redfont{0.022} & 0.260 & 11.819   \\
    {wo $L_{con}$}   & 0.458 & \bluefont{0.024} & 0.255 & \redfont{10.376}   \\
    {wo $L_{vel}$}   & \redfont{0.469} & 0.026 & \redfont{0.229} & 10.758   \\
    {Ours (full)}    & \bluefont{0.461} & 0.026 & \bluefont{0.237} & \bluefont{10.713}   \\
    \hline
    \end{tabular}
    }
    \caption{Ablation study of loss functions. The best performance is highlighted in \redfont{red} (1st best) and \bluefont{blue} (2nd best).}
    \label{tab:ablation:loss_function}
\end{table}

% *********************************************************
\subsection{Ablation Study}

To assess the impact of different components of our method associated with loss items, we remove each term in the loss function and evaluate the influence in the cross-reenactment experiment. The quantitative results are shown in \cref{tab:ablation:loss_function}. It can be observed that the adversarial loss $L_{gan}$ and the perceptual loss $L_{per}$ play major roles in the generation process as their removal causes an overall degradation in the generation quality. In addition, the consistency loss $L_{con}$ helps to keep the face parameters consistent between the generated and driven images. Finally, the removal of the velocity loss $L_{vel}$ results in a slightly worse quality, which is due to the correlation of facial details between adjacent frames of a video. In general, our method exhibits superior performance benefiting from well-designed loss functions.

% Both FID and CSIM show dramatic performance degradation after removing adversarial loss $L_{gan}$. This is because the adversarial loss forces the generated images to be similar to the real images, which helps to improve the generation quality. There is an increase in FID after removing perceptual loss $L_{per}$. This is because the VGG network captures image texture, which helps constrain the texture similarity between the real and generated images. After removing the consistency loss $L_{con}$, AED shows an increase. The reason is that the consistency loss helps to keep the face parameters consistent between the generated and driven images. The FID increases after removing the velocity loss $L_{vel}$, which is due to the correlation of facial details between adjacent frames of a video.

In order to assess the impact of the size of the face parameter space provided by the morphable prior, we gradually increase the number of videos $k$ in the auxiliary data and test our method in a cross-reenactment experiment. Considering the inability to accurately measure the difference in face parameter diversity between videos, we chose to add one video at a time to the auxiliary data. The quantitative results are shown in \cref{tab:ablation:data_size}. It can be observed that the generation quality improves as the size of auxiliary data increases. This suggests that the full morphable prior space can be approached as the size of the auxiliary data increases.

% *********************************************************
\subsection{Feature Visualisation}
\begin{figure}[t]
    \centering
    \includegraphics[width = 0.47\textwidth]{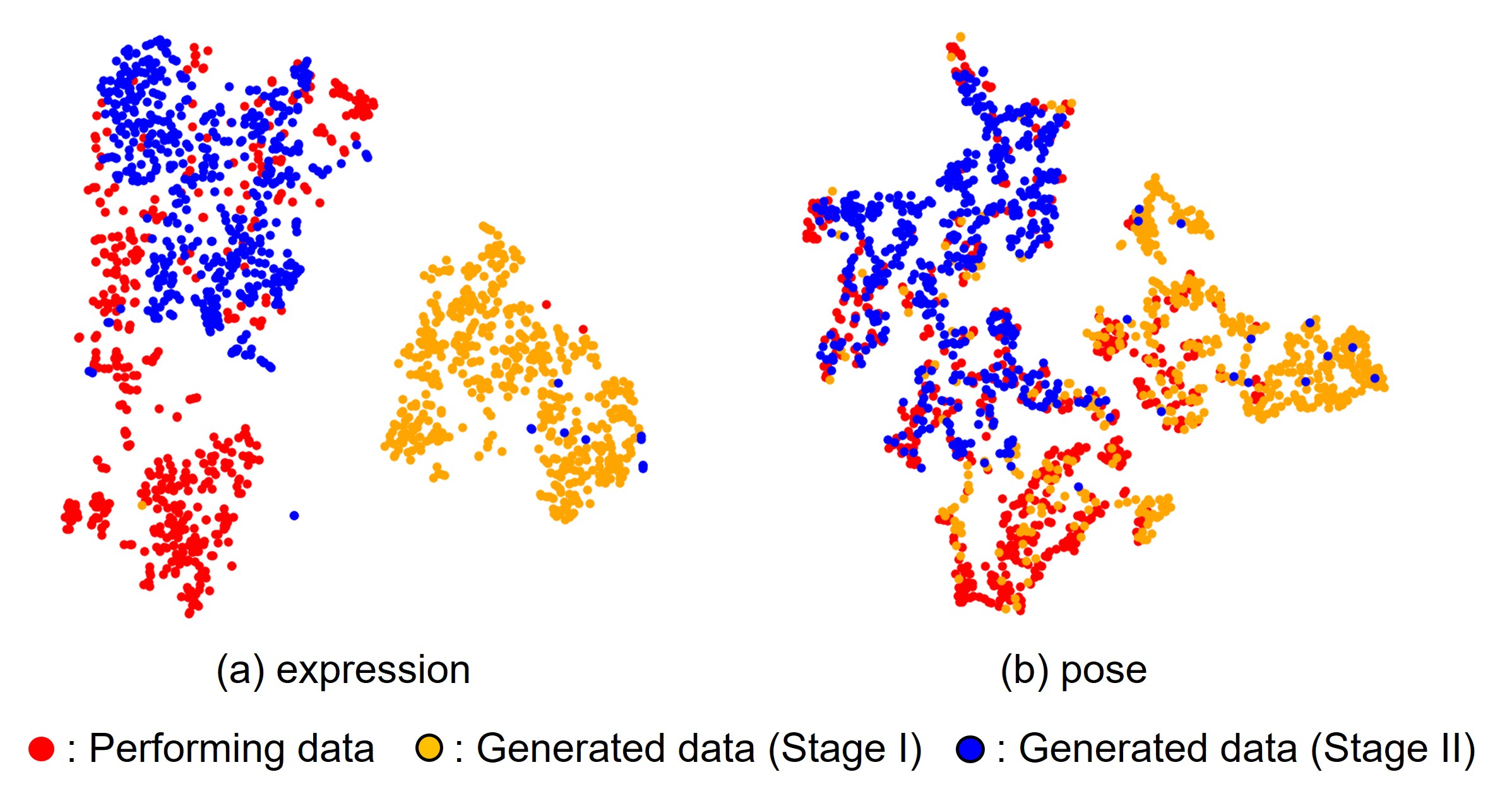}
    \caption{Visualisation of face parameters before and after the scalable training stage.}
    \label{img:tsne}
\end{figure}

In order to compare the structural differences of the face parameters before and after the scalable training stage, we use t-SNE \cite{van2008visualizing} to visualize the expression and pose coefficients, respectively. There are three categories of face parameters shown in \cref{img:tsne}, the monocular video used for training (Performing data, red dots), the face parameters corresponding to stage I (yellow dots), and the face parameters corresponding to the images generated by stage II (blue dots). It can be observed that in both \cref{img:tsne} (a) and \cref{img:tsne} (b), the yellow dots are far away from the red dots, while the blue dots are entangled with the red dots. The driven data in stage II are forced to align with the performing data. This suggests that our training strategy, by introducing the auxiliary data, helps to mitigate the effects caused by the different distribution between the training and testing data. In addition to the personalized prior provided by a monocular video, we introduce the morphable prior to enhance the expressiveness of the trained model, resulting in high-quality personalized details under various semantic parameters. 

% T-SNE \cite{van2008visualizing} is a dimensionality reduction technique suitable for visualizing high-dimensional features in datasets. 

% It should be noted that there are differences in the 3DMM neutral templates corresponding to different identities. Even if the 3DMM coefficients corresponding to the generated images are consistent with the driven data, it is not guaranteed that the expressions and poses of the generated images are exactly identical to those in the driven images. Therefore, instead of excessively pursuing such consistency, our method aligns the generated images with the performance images at the expense of certain face parameter consistency, effectively improving the generation quality. 

% This means that our method can suppress the effects of differences in 3DMM neutral templates between different identities by introducing a rich morphable prior.

\begin{table}[t]
    \centering
    \resizebox{0.45\textwidth}{!}{
    \begin{tabular}{cccccccc} 
    \hline
    Method & CSIM($\uparrow$) & APD($\downarrow$) & AED($\downarrow$) & FID($\downarrow$)   \\ 
    \hline
    {Ours ($k=0$)}  & 0.403 & 0.028 & 0.245 & 23.700  \\
    % \hline
    {Ours ($k=1$)}   & 0.441 & 0.029 & 0.325 & 12.652   \\
    {Ours ($k=2$)}   & 0.456 & 0.027 & 0.247 & 11.156   \\
    {Ours ($k=3$)}    & \textbf{0.461} & \textbf{0.026} & \textbf{0.237} & \textbf{10.713}   \\
    \hline
    \end{tabular}
    }
    \caption{Ablation study for gradually adding the number of videos in the auxiliary data. The bold results are the best.}
    \label{tab:ablation:data_size}
\end{table}

%-------------------------------------------------------------------------

\section{Conclusion and Discussion}
\label{sec:con}
%-------------------------------------------------------------------------

\noindent \textbf{Conclusion.} In this work, we present MyPortrait, a simple, general, and flexible framework for neural portrait generation. By exploiting personalized prior from a monocular video and morphable prior provided by 3DMM, our framework generates high-quality personalized dynamic faces. Comprehensive experiments demonstrate that our method outperforms the state-of-the-art ones. This work provides a novel perspective in generating personalized details for future research on face reenactment.

\noindent \textbf{Limitation.} Benefiting from the reduction of 3D to 2D scenes, our method can be trained using a simple network structure. However, this also results in that our method can only be applied to monocular videos with fixed backgrounds. It is promising to combine our method with face segmentation methods in the future. In addition, the accuracy of face parameters extracted by face trackers greatly influences our method's performance. With the development of face trackers, the performance of our method will be further improved.

{
    \small
    \bibliographystyle{ieeenat_fullname}
    \bibliography{main}

\begin{thebibliography}{65}
\providecommand{\natexlab}[1]{#1}
\providecommand{\url}[1]{\texttt{#1}}
\expandafter\ifx\csname urlstyle\endcsname\relax
  \providecommand{\doi}[1]{doi: #1}\else
  \providecommand{\doi}{doi: \begingroup \urlstyle{rm}\Url}\fi

\bibitem[Amodei et~al.(2016)Amodei, Ananthanarayanan, Anubhai, Bai, Battenberg, Case, Casper, Catanzaro, Cheng, Chen, et~al.]{amodei2016deep}
Dario Amodei, Sundaram Ananthanarayanan, Rishita Anubhai, Jingliang Bai, Eric Battenberg, Carl Case, Jared Casper, Bryan Catanzaro, Qiang Cheng, Guoliang Chen, et~al.
\newblock Deep speech 2: End-to-end speech recognition in english and mandarin.
\newblock In \emph{International Conference on Machine Learning}, pages 173--182, 2016.

\bibitem[Bai et~al.(2023)Bai, Fan, Wang, Zhang, Sun, Yuan, and Shan]{bai2023high}
Yunpeng Bai, Yanbo Fan, Xuan Wang, Yong Zhang, Jingxiang Sun, Chun Yuan, and Ying Shan.
\newblock High-fidelity facial avatar reconstruction from monocular video with generative priors.
\newblock In \emph{Proceedings of the IEEE/CVF Conference on Computer Vision and Pattern Recognition}, pages 4541--4551, 2023.

\bibitem[Blanz and Vetter(1999)]{blanz1999morphable}
Volker Blanz and Thomas Vetter.
\newblock A morphable model for the synthesis of 3d faces.
\newblock In \emph{Proceedings of the 26th annual conference on Computer graphics and interactive techniques}, pages 187--194, 1999.

\bibitem[Booth et~al.(2016)Booth, Roussos, Zafeiriou, Ponniah, and Dunaway]{booth20163d}
James Booth, Anastasios Roussos, Stefanos Zafeiriou, Allan Ponniah, and David Dunaway.
\newblock A 3d morphable model learnt from 10,000 faces.
\newblock In \emph{Proceedings of the IEEE conference on Computer Vision and Pattern Recognition}, pages 5543--5552, 2016.

\bibitem[Chan et~al.(2022)Chan, Lin, Chan, Nagano, Pan, De~Mello, Gallo, Guibas, Tremblay, Khamis, et~al.]{chan2022efficient}
Eric~R Chan, Connor~Z Lin, Matthew~A Chan, Koki Nagano, Boxiao Pan, Shalini De~Mello, Orazio Gallo, Leonidas~J Guibas, Jonathan Tremblay, Sameh Khamis, et~al.
\newblock Efficient geometry-aware 3d generative adversarial networks.
\newblock In \emph{Proceedings of the IEEE/CVF Conference on Computer Vision and Pattern Recognition}, pages 16123--16133, 2022.

\bibitem[Deng et~al.(2019)Deng, Guo, Xue, and Zafeiriou]{deng2019arcface}
Jiankang Deng, Jia Guo, Niannan Xue, and Stefanos Zafeiriou.
\newblock Arcface: Additive angular margin loss for deep face recognition.
\newblock In \emph{Proceedings of the IEEE/CVF Conference on Computer Vision and Pattern Recognition}, pages 4690--4699, 2019.

\bibitem[Dhere et~al.(2020)Dhere, Rathod, Aarankalle, Lad, and Gandhi]{dhere2020review}
Sourabh Dhere, Suresh~B Rathod, Sanket Aarankalle, Yash Lad, and Megh Gandhi.
\newblock A review on face reenactment techniques.
\newblock In \emph{2020 International Conference on Industry 4.0 Technology}, pages 191--194. IEEE, 2020.

\bibitem[Doukas et~al.(2021)Doukas, Koujan, Sharmanska, Roussos, and Zafeiriou]{doukas2021head2head++}
Michail~Christos Doukas, Mohammad~Rami Koujan, Viktoriia Sharmanska, Anastasios Roussos, and Stefanos Zafeiriou.
\newblock Head2head++: Deep facial attributes re-targeting.
\newblock \emph{IEEE Transactions on Biometrics, Behavior, and Identity Science}, 3\penalty0 (1):\penalty0 31--43, 2021.

\bibitem[Doukas et~al.(2023)Doukas, Ploumpis, and Zafeiriou]{doukas2023dynamic}
Michail~Christos Doukas, Stylianos Ploumpis, and Stefanos Zafeiriou.
\newblock Dynamic neural portraits.
\newblock In \emph{Proceedings of the IEEE/CVF Winter Conference on Applications of Computer Vision}, pages 4073--4083, 2023.

\bibitem[Feng et~al.(2021)Feng, Feng, Black, and Bolkart]{feng2021learning}
Yao Feng, Haiwen Feng, Michael~J Black, and Timo Bolkart.
\newblock Learning an animatable detailed 3d face model from in-the-wild images.
\newblock \emph{ACM Transactions on Graphics}, 40\penalty0 (4):\penalty0 1--13, 2021.

\bibitem[Gafni et~al.(2021)Gafni, Thies, Zollhofer, and Nie{\ss}ner]{gafni2021dynamic}
Guy Gafni, Justus Thies, Michael Zollhofer, and Matthias Nie{\ss}ner.
\newblock Dynamic neural radiance fields for monocular 4d facial avatar reconstruction.
\newblock In \emph{Proceedings of the IEEE/CVF Conference on Computer Vision and Pattern Recognition}, pages 8649--8658, 2021.

\bibitem[Gandelsman et~al.(2022)Gandelsman, Sun, Chen, and Efros]{gandelsman2022test}
Yossi Gandelsman, Yu Sun, Xinlei Chen, and Alexei Efros.
\newblock Test-time training with masked autoencoders.
\newblock \emph{Advances in Neural Information Processing Systems}, 35:\penalty0 29374--29385, 2022.

\bibitem[Goodfellow et~al.(2014)Goodfellow, Pouget-Abadie, Mirza, Xu, Warde-Farley, Ozair, Courville, and Bengio]{goodfellow2014generative}
Ian Goodfellow, Jean Pouget-Abadie, Mehdi Mirza, Bing Xu, David Warde-Farley, Sherjil Ozair, Aaron Courville, and Yoshua Bengio.
\newblock Generative adversarial nets.
\newblock \emph{Advances in neural information processing systems}, 27, 2014.

\bibitem[Grassal et~al.(2022)Grassal, Prinzler, Leistner, Rother, Nie{\ss}ner, and Thies]{grassal2022neural}
Philip-William Grassal, Malte Prinzler, Titus Leistner, Carsten Rother, Matthias Nie{\ss}ner, and Justus Thies.
\newblock Neural head avatars from monocular rgb videos.
\newblock In \emph{Proceedings of the IEEE/CVF Conference on Computer Vision and Pattern Recognition}, pages 18653--18664, 2022.

\bibitem[Guo et~al.(2021)Guo, Chen, Liang, Liu, Bao, and Zhang]{guo2021ad}
Yudong Guo, Keyu Chen, Sen Liang, Yong-Jin Liu, Hujun Bao, and Juyong Zhang.
\newblock Ad-nerf: Audio driven neural radiance fields for talking head synthesis.
\newblock In \emph{Proceedings of the IEEE/CVF International Conference on Computer Vision}, pages 5784--5794, 2021.

\bibitem[Hannun et~al.(2014)Hannun, Case, Casper, Catanzaro, Diamos, Elsen, Prenger, Satheesh, Sengupta, Coates, et~al.]{hannun2014deep}
Awni Hannun, Carl Case, Jared Casper, Bryan Catanzaro, Greg Diamos, Erich Elsen, Ryan Prenger, Sanjeev Satheesh, Shubho Sengupta, Adam Coates, et~al.
\newblock Deep speech: Scaling up end-to-end speech recognition.
\newblock \emph{arXiv preprint arXiv:1412.5567}, 2014.

\bibitem[Heusel et~al.(2017)Heusel, Ramsauer, Unterthiner, Nessler, and Hochreiter]{heusel2017gans}
Martin Heusel, Hubert Ramsauer, Thomas Unterthiner, Bernhard Nessler, and Sepp Hochreiter.
\newblock Gans trained by a two time-scale update rule converge to a local nash equilibrium.
\newblock \emph{Advances in neural information processing systems}, 30, 2017.

\bibitem[Hong et~al.(2022)Hong, Zhang, Shen, and Xu]{hong2022depth}
Fa-Ting Hong, Longhao Zhang, Li Shen, and Dan Xu.
\newblock Depth-aware generative adversarial network for talking head video generation.
\newblock In \emph{Proceedings of the IEEE/CVF Conference on Computer Vision and Pattern Recognition}, pages 3397--3406, 2022.

\bibitem[Isola et~al.(2017)Isola, Zhu, Zhou, and Efros]{isola2017image}
Phillip Isola, Jun-Yan Zhu, Tinghui Zhou, and Alexei~A Efros.
\newblock Image-to-image translation with conditional adversarial networks.
\newblock In \emph{Proceedings of the IEEE/CVF Conference on Computer Vision and Pattern Recognition}, pages 1125--1134, 2017.

\bibitem[Johnson et~al.(2016)Johnson, Alahi, and Fei-Fei]{johnson2016perceptual}
Justin Johnson, Alexandre Alahi, and Li Fei-Fei.
\newblock Perceptual losses for real-time style transfer and super-resolution.
\newblock In \emph{Proceedings of the European Conference on Computer Vision}, pages 694--711. Springer, 2016.

\bibitem[Karras et~al.(2019)Karras, Laine, and Aila]{karras2019style}
Tero Karras, Samuli Laine, and Timo Aila.
\newblock A style-based generator architecture for generative adversarial networks.
\newblock In \emph{Proceedings of the IEEE/CVF Conference on Computer Vision and Pattern Recognition}, pages 4401--4410, 2019.

\bibitem[Karras et~al.(2020)Karras, Laine, Aittala, Hellsten, Lehtinen, and Aila]{karras2020analyzing}
Tero Karras, Samuli Laine, Miika Aittala, Janne Hellsten, Jaakko Lehtinen, and Timo Aila.
\newblock Analyzing and improving the image quality of stylegan.
\newblock In \emph{Proceedings of the IEEE/CVF Conference on Computer Vision and Pattern Recognition}, pages 8110--8119, 2020.

\bibitem[Karras et~al.(2021)Karras, Aittala, Laine, H{\"a}rk{\"o}nen, Hellsten, Lehtinen, and Aila]{karras2021alias}
Tero Karras, Miika Aittala, Samuli Laine, Erik H{\"a}rk{\"o}nen, Janne Hellsten, Jaakko Lehtinen, and Timo Aila.
\newblock Alias-free generative adversarial networks.
\newblock \emph{Advances in Neural Information Processing Systems}, 34:\penalty0 852--863, 2021.

\bibitem[Kellnhofer et~al.(2019)Kellnhofer, Recasens, Stent, Matusik, and Torralba]{kellnhofer2019gaze360}
Petr Kellnhofer, Adria Recasens, Simon Stent, Wojciech Matusik, and Antonio Torralba.
\newblock Gaze360: Physically unconstrained gaze estimation in the wild.
\newblock In \emph{Proceedings of the IEEE/CVF International Conference on Computer Vision}, pages 6912--6921, 2019.

\bibitem[Kim et~al.(2018)Kim, Garrido, Tewari, Xu, Thies, Niessner, P{\'e}rez, Richardt, Zollh{\"o}fer, and Theobalt]{kim2018deep}
Hyeongwoo Kim, Pablo Garrido, Ayush Tewari, Weipeng Xu, Justus Thies, Matthias Niessner, Patrick P{\'e}rez, Christian Richardt, Michael Zollh{\"o}fer, and Christian Theobalt.
\newblock Deep video portraits.
\newblock \emph{ACM Transactions on Graphics}, 37\penalty0 (4):\penalty0 1--14, 2018.

\bibitem[Kingma and Ba(2015)]{kingma2014adam}
Diederik~P Kingma and Jimmy Ba.
\newblock Adam: A method for stochastic optimization.
\newblock In \emph{International Conference on Learning Representations}, 2015.

\bibitem[Koujan et~al.(2020)Koujan, Doukas, Roussos, and Zafeiriou]{koujan2020head2head}
Mohammad~Rami Koujan, Michail~Christos Doukas, Anastasios Roussos, and Stefanos Zafeiriou.
\newblock Head2head: Video-based neural head synthesis.
\newblock In \emph{2020 15th IEEE International Conference on Automatic Face and Gesture Recognition}, pages 16--23. IEEE, 2020.

\bibitem[Li et~al.(2017)Li, Bolkart, Black, Li, and Romero]{li2017learning}
Tianye Li, Timo Bolkart, Michael~J Black, Hao Li, and Javier Romero.
\newblock Learning a model of facial shape and expression from 4d scans.
\newblock \emph{ACM Trans. Graph.}, 36\penalty0 (6):\penalty0 194--1, 2017.

\bibitem[Ma et~al.(2023)Ma, Zhu, Qi, Lei, and Zhang]{ma2023otavatar}
Zhiyuan Ma, Xiangyu Zhu, Guo-Jun Qi, Zhen Lei, and Lei Zhang.
\newblock Otavatar: One-shot talking face avatar with controllable tri-plane rendering.
\newblock In \emph{Proceedings of the IEEE/CVF Conference on Computer Vision and Pattern Recognition}, pages 16901--16910, 2023.

\bibitem[Medin et~al.(2022)Medin, Egger, Cherian, Wang, Tenenbaum, Liu, and Marks]{medin2022most}
Safa~C Medin, Bernhard Egger, Anoop Cherian, Ye Wang, Joshua~B Tenenbaum, Xiaoming Liu, and Tim~K Marks.
\newblock Most-gan: 3d morphable stylegan for disentangled face image manipulation.
\newblock In \emph{Proceedings of the AAAI Conference on Artificial Intelligence}, pages 1962--1971, 2022.

\bibitem[Miyato et~al.(2018)Miyato, Kataoka, Koyama, and Yoshida]{miyato2018spectral}
Takeru Miyato, Toshiki Kataoka, Masanori Koyama, and Yuichi Yoshida.
\newblock Spectral normalization for generative adversarial networks.
\newblock In \emph{International Conference on Learning Representations}, 2018.

\bibitem[Nitzan et~al.(2022)Nitzan, Aberman, He, Liba, Yarom, Gandelsman, Mosseri, Pritch, and Cohen-Or]{nitzan2022mystyle}
Yotam Nitzan, Kfir Aberman, Qiurui He, Orly Liba, Michal Yarom, Yossi Gandelsman, Inbar Mosseri, Yael Pritch, and Daniel Cohen-Or.
\newblock Mystyle: A personalized generative prior.
\newblock \emph{ACM Transactions on Graphics}, 41\penalty0 (6):\penalty0 1--10, 2022.

\bibitem[Odena et~al.(2016)Odena, Dumoulin, and Olah]{odena2016deconvolution}
Augustus Odena, Vincent Dumoulin, and Chris Olah.
\newblock Deconvolution and checkerboard artifacts.
\newblock \emph{Distill}, 1\penalty0 (10):\penalty0 e3, 2016.

\bibitem[Oorloff and Yacoob(2023)]{oorloff2023robust}
Trevine Oorloff and Yaser Yacoob.
\newblock Robust one-shot face video re-enactment using hybrid latent spaces of stylegan2.
\newblock In \emph{Proceedings of the IEEE/CVF International Conference on Computer Vision}, pages 20947--20957, 2023.

\bibitem[Paszke et~al.(2019)Paszke, Gross, Massa, Lerer, Bradbury, Chanan, Killeen, Lin, Gimelshein, Antiga, et~al.]{paszke2019pytorch}
Adam Paszke, Sam Gross, Francisco Massa, Adam Lerer, James Bradbury, Gregory Chanan, Trevor Killeen, Zeming Lin, Natalia Gimelshein, Luca Antiga, et~al.
\newblock Pytorch: An imperative style, high-performance deep learning library.
\newblock \emph{Advances in Neural Information Processing Systems}, 32, 2019.

\bibitem[Paysan et~al.(2009)Paysan, Knothe, Amberg, Romdhani, and Vetter]{paysan20093d}
Pascal Paysan, Reinhard Knothe, Brian Amberg, Sami Romdhani, and Thomas Vetter.
\newblock A 3d face model for pose and illumination invariant face recognition.
\newblock In \emph{2009 sixth IEEE international conference on advanced video and signal based surveillance}, pages 296--301. Ieee, 2009.

\bibitem[Ren et~al.(2021)Ren, Li, Chen, Li, and Liu]{ren2021pirenderer}
Yurui Ren, Ge Li, Yuanqi Chen, Thomas~H Li, and Shan Liu.
\newblock Pirenderer: Controllable portrait image generation via semantic neural rendering.
\newblock In \emph{Proceedings of the IEEE/CVF International Conference on Computer Vision}, pages 13759--13768, 2021.

\bibitem[Sha et~al.(2023)Sha, Zhang, Shen, Li, and Mei]{sha2023deep}
Tong Sha, Wei Zhang, Tong Shen, Zhoujun Li, and Tao Mei.
\newblock Deep person generation: A survey from the perspective of face, pose, and cloth synthesis.
\newblock \emph{ACM Computing Surveys}, 55\penalty0 (12):\penalty0 1--37, 2023.

\bibitem[Siarohin et~al.(2019{\natexlab{a}})Siarohin, Lathuili{\`e}re, Tulyakov, Ricci, and Sebe]{siarohin2019animating}
Aliaksandr Siarohin, St{\'e}phane Lathuili{\`e}re, Sergey Tulyakov, Elisa Ricci, and Nicu Sebe.
\newblock Animating arbitrary objects via deep motion transfer.
\newblock In \emph{Proceedings of the IEEE/CVF Conference on Computer Vision and Pattern Recognition}, pages 2377--2386, 2019{\natexlab{a}}.

\bibitem[Siarohin et~al.(2019{\natexlab{b}})Siarohin, Lathuili{\`e}re, Tulyakov, Ricci, and Sebe]{siarohin2019first}
Aliaksandr Siarohin, St{\'e}phane Lathuili{\`e}re, Sergey Tulyakov, Elisa Ricci, and Nicu Sebe.
\newblock First order motion model for image animation.
\newblock In \emph{Proceedings of the 33rd International Conference on Neural Information Processing Systems}, pages 7137--7147, 2019{\natexlab{b}}.

\bibitem[Siarohin et~al.(2021)Siarohin, Woodford, Ren, Chai, and Tulyakov]{siarohin2021motion}
Aliaksandr Siarohin, Oliver~J Woodford, Jian Ren, Menglei Chai, and Sergey Tulyakov.
\newblock Motion representations for articulated animation.
\newblock In \emph{Proceedings of the IEEE/CVF Conference on Computer Vision and Pattern Recognition}, pages 13653--13662, 2021.

\bibitem[Simonyan and Zisserman(2015)]{simonyan2014very}
Karen Simonyan and Andrew Zisserman.
\newblock Very deep convolutional networks for large-scale image recognition.
\newblock In \emph{International Conference on Learning Representations}, 2015.

\bibitem[Sun et~al.(2023{\natexlab{a}})Sun, Wang, Wang, Li, Zhang, Zhang, and Liu]{sun2023next3d}
Jingxiang Sun, Xuan Wang, Lizhen Wang, Xiaoyu Li, Yong Zhang, Hongwen Zhang, and Yebin Liu.
\newblock Next3d: Generative neural texture rasterization for 3d-aware head avatars.
\newblock In \emph{Proceedings of the IEEE/CVF Conference on Computer Vision and Pattern Recognition}, pages 20991--21002, 2023{\natexlab{a}}.

\bibitem[Sun et~al.(2023{\natexlab{b}})Sun, Li, Dalal, Hsu, Koyejo, Guestrin, Wang, Hashimoto, and Chen]{sun2023learning}
Yu Sun, Xinhao Li, Karan Dalal, Chloe Hsu, Sanmi Koyejo, Carlos Guestrin, Xiaolong Wang, Tatsunori Hashimoto, and Xinlei Chen.
\newblock Learning to (learn at test time).
\newblock \emph{arXiv preprint arXiv:2310.13807}, 2023{\natexlab{b}}.

\bibitem[Van~der Maaten and Hinton(2008)]{van2008visualizing}
Laurens Van~der Maaten and Geoffrey Hinton.
\newblock Visualizing data using t-sne.
\newblock \emph{Journal of Machine Learning Research}, 9\penalty0 (11):\penalty0 2579--2605, 2008.

\bibitem[Wang et~al.(2023{\natexlab{a}})Wang, Liang, Zhou, Tang, Wu, He, Hong, Liu, Ding, Liu, et~al.]{wang2023robust}
Kaisiyuan Wang, Changcheng Liang, Hang Zhou, Jiaxiang Tang, Qianyi Wu, Dongliang He, Zhibin Hong, Jingtuo Liu, Errui Ding, Ziwei Liu, et~al.
\newblock Robust video portrait reenactment via personalized representation quantization.
\newblock In \emph{Proceedings of the AAAI Conference on Artificial Intelligence}, pages 2564--2572, 2023{\natexlab{a}}.

\bibitem[Wang et~al.(2023{\natexlab{b}})Wang, Zhou, Wu, Tang, Xu, Liang, Hu, Ding, Liu, Liu, et~al.]{wang2023efficient}
Kaisiyuan Wang, Hang Zhou, Qianyi Wu, Jiaxiang Tang, Zhiliang Xu, Borong Liang, Tianshu Hu, Errui Ding, Jingtuo Liu, Ziwei Liu, et~al.
\newblock Efficient video portrait reenactment via grid-based codebook.
\newblock In \emph{ACM SIGGRAPH 2023 Conference Proceedings}, pages 1--9, 2023{\natexlab{b}}.

\bibitem[Wang et~al.(2023{\natexlab{c}})Wang, Zhao, Sun, Zhang, Zhang, Yu, and Liu]{wang2023styleavatar}
Lizhen Wang, Xiaochen Zhao, Jingxiang Sun, Yuxiang Zhang, Hongwen Zhang, Tao Yu, and Yebin Liu.
\newblock Styleavatar: Real-time photo-realistic portrait avatar from a single video.
\newblock In \emph{ACM SIGGRAPH 2023 Conference Proceedings}, pages 67:1--67:10, 2023{\natexlab{c}}.

\bibitem[Wang et~al.(2022)Wang, Yang, Bremond, and Dantcheva]{wang2022latent}
Yaohui Wang, Di Yang, Francois Bremond, and Antitza Dantcheva.
\newblock Latent image animator: Learning to animate images via latent space navigation.
\newblock In \emph{The International Conference on Learning Representations}, 2022.

\bibitem[Wiles et~al.(2018)Wiles, Koepke, and Zisserman]{wiles2018x2face}
Olivia Wiles, A Koepke, and Andrew Zisserman.
\newblock X2face: A network for controlling face generation using images, audio, and pose codes.
\newblock In \emph{Proceedings of the European Conference on Computer Vision}, pages 670--686, 2018.

\bibitem[Xing et~al.(2023)Xing, Xia, Zhang, Cun, Wang, and Wong]{xing2023codetalker}
Jinbo Xing, Menghan Xia, Yuechen Zhang, Xiaodong Cun, Jue Wang, and Tien-Tsin Wong.
\newblock Codetalker: Speech-driven 3d facial animation with discrete motion prior.
\newblock In \emph{Proceedings of the IEEE/CVF Conference on Computer Vision and Pattern Recognition}, pages 12780--12790, 2023.

\bibitem[Xu et~al.(2022)Xu, Zhang, Han, Tian, Zeng, Tai, Wang, Wang, and Liu]{xu2022designing}
Chao Xu, Jiangning Zhang, Yue Han, Guanzhong Tian, Xianfang Zeng, Ying Tai, Yabiao Wang, Chengjie Wang, and Yong Liu.
\newblock Designing one unified framework for high-fidelity face reenactment and swapping.
\newblock In \emph{European Conference on Computer Vision}, pages 54--71. Springer, 2022.

\bibitem[Xu et~al.(2023)Xu, Zhang, Wang, Zhao, Huang, Qi, and Liu]{xu2023latentavatar}
Yuelang Xu, Hongwen Zhang, Lizhen Wang, Xiaochen Zhao, Han Huang, Guojun Qi, and Yebin Liu.
\newblock Latentavatar: Learning latent expression code for expressive neural head avatar.
\newblock In \emph{ACM SIGGRAPH 2023 Conference Proceedings}, pages 86:1--86:10, 2023.

\bibitem[Yang et~al.(2022)Yang, Chen, Guo, Zhang, Guo, and Zhang]{yang2022face2face}
Kewei Yang, Kang Chen, Daoliang Guo, Song-Hai Zhang, Yuan-Chen Guo, and Weidong Zhang.
\newblock Face2face $\rho$: Real-time high-resolution one-shot face reenactment.
\newblock In \emph{European Conference on Computer Vision}, pages 55--71. Springer, 2022.

\bibitem[Yin et~al.(2022)Yin, Zhang, Cun, Cao, Fan, Wang, Bai, Wu, Wang, and Yang]{yin2022styleheat}
Fei Yin, Yong Zhang, Xiaodong Cun, Mingdeng Cao, Yanbo Fan, Xuan Wang, Qingyan Bai, Baoyuan Wu, Jue Wang, and Yujiu Yang.
\newblock Styleheat: One-shot high-resolution editable talking face generation via pre-trained stylegan.
\newblock In \emph{Proceedings of the European Conference on Computer Vision}, pages 85--101. Springer, 2022.

\bibitem[Zeng et~al.(2023)Zeng, Chen, Xu, and Kalantari]{zeng2023mystyle++}
Libing Zeng, Lele Chen, Yi Xu, and Nima Kalantari.
\newblock Mystyle++: A controllable personalized generative prior.
\newblock \emph{arXiv preprint arXiv:2306.04865}, 2023.

\bibitem[Zhan et~al.(2023)Zhan, Yu, Wu, Zhang, Lu, Liu, Kortylewski, Theobalt, and Xing]{zhan2023multimodal}
Fangneng Zhan, Yingchen Yu, Rongliang Wu, Jiahui Zhang, Shijian Lu, Lingjie Liu, Adam Kortylewski, Christian Theobalt, and Eric Xing.
\newblock Multimodal image synthesis and editing: A survey and taxonomy.
\newblock \emph{IEEE Transactions on Pattern Analysis and Machine Intelligence}, 2023.

\bibitem[Zhang et~al.(2023{\natexlab{a}})Zhang, Qi, Zhang, Zhang, Wu, Chen, Chen, Wang, and Wen]{zhang2023metaportrait}
Bowen Zhang, Chenyang Qi, Pan Zhang, Bo Zhang, HsiangTao Wu, Dong Chen, Qifeng Chen, Yong Wang, and Fang Wen.
\newblock Metaportrait: Identity-preserving talking head generation with fast personalized adaptation.
\newblock In \emph{Proceedings of the IEEE/CVF Conference on Computer Vision and Pattern Recognition}, pages 22096--22105, 2023{\natexlab{a}}.

\bibitem[Zhang et~al.(2019)Zhang, Goodfellow, Metaxas, and Odena]{zhang2019self}
Han Zhang, Ian Goodfellow, Dimitris Metaxas, and Augustus Odena.
\newblock Self-attention generative adversarial networks.
\newblock In \emph{International Conference on Machine Learning}, pages 7354--7363. PMLR, 2019.

\bibitem[Zhang et~al.(2018)Zhang, Isola, Efros, Shechtman, and Wang]{zhang2018unreasonable}
Richard Zhang, Phillip Isola, Alexei~A Efros, Eli Shechtman, and Oliver Wang.
\newblock The unreasonable effectiveness of deep features as a perceptual metric.
\newblock In \emph{Proceedings of the IEEE Conference on Computer Vision and Pattern Recognition}, pages 586--595, 2018.

\bibitem[Zhang et~al.(2023{\natexlab{b}})Zhang, Cun, Wang, Zhang, Shen, Guo, Shan, and Wang]{zhang2023sadtalker}
Wenxuan Zhang, Xiaodong Cun, Xuan Wang, Yong Zhang, Xi Shen, Yu Guo, Ying Shan, and Fei Wang.
\newblock Sadtalker: Learning realistic 3d motion coefficients for stylized audio-driven single image talking face animation.
\newblock In \emph{Proceedings of the IEEE/CVF Conference on Computer Vision and Pattern Recognition}, pages 8652--8661, 2023{\natexlab{b}}.

\bibitem[Zhao and Zhang(2022)]{zhao2022thin}
Jian Zhao and Hui Zhang.
\newblock Thin-plate spline motion model for image animation.
\newblock In \emph{Proceedings of the IEEE/CVF Conference on Computer Vision and Pattern Recognition}, pages 3657--3666, 2022.

\bibitem[Zhen et~al.(2023)Zhen, Song, He, Cao, Shi, and Luo]{zhen2023human}
Rui Zhen, Wenchao Song, Qiang He, Juan Cao, Lei Shi, and Jia Luo.
\newblock Human-computer interaction system: A survey of talking-head generation.
\newblock \emph{Electronics}, 12\penalty0 (1):\penalty0 218, 2023.

\bibitem[Zheng et~al.(2022)Zheng, Abrevaya, B{\"u}hler, Chen, Black, and Hilliges]{zheng2022avatar}
Yufeng Zheng, Victoria~Fern{\'a}ndez Abrevaya, Marcel~C B{\"u}hler, Xu Chen, Michael~J Black, and Otmar Hilliges.
\newblock Im avatar: Implicit morphable head avatars from videos.
\newblock In \emph{Proceedings of the IEEE/CVF Conference on Computer Vision and Pattern Recognition}, pages 13545--13555, 2022.

\bibitem[Zhu et~al.(2017)Zhu, Park, Isola, and Efros]{zhu2017unpaired}
Jun-Yan Zhu, Taesung Park, Phillip Isola, and Alexei~A Efros.
\newblock Unpaired image-to-image translation using cycle-consistent adversarial networks.
\newblock In \emph{Proceedings of the IEEE International Conference on Computer Vision}, pages 2223--2232, 2017.

\end{thebibliography}
}
% *****************************************************
% supplementary material
\clearpage
\setcounter{page}{1}
\maketitlesupplementary

%  ******************************************************
\section{Dataset}
\label{sec:dataset}

\begin{figure}[t]
    \centering
    \subfloat[ID.1]{
        \includegraphics[width =0.1\textwidth]
        {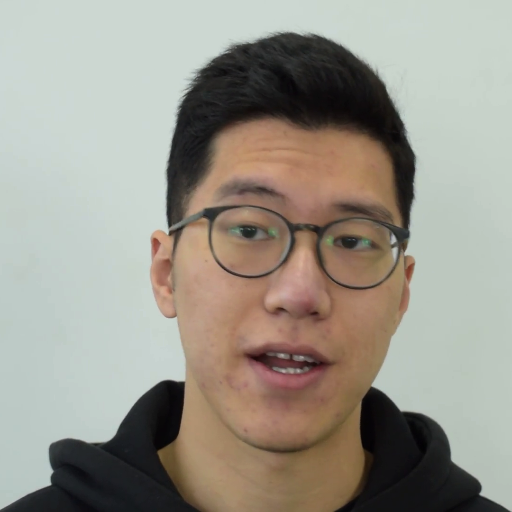}
        }
    \subfloat[ID.2]{
        \includegraphics[width =0.1\textwidth]
        {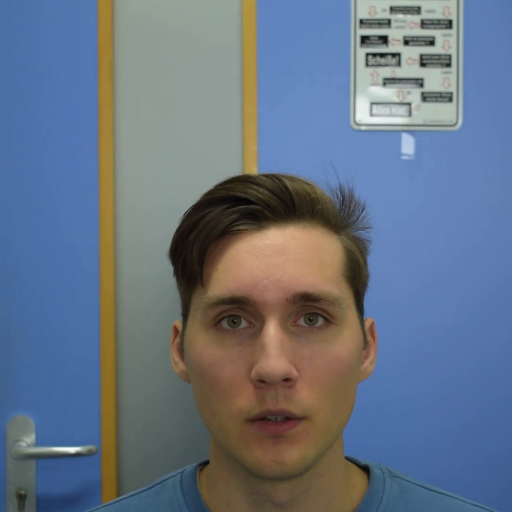}
        }
    \subfloat[ID.3]{
        \includegraphics[width =0.1\textwidth]
        {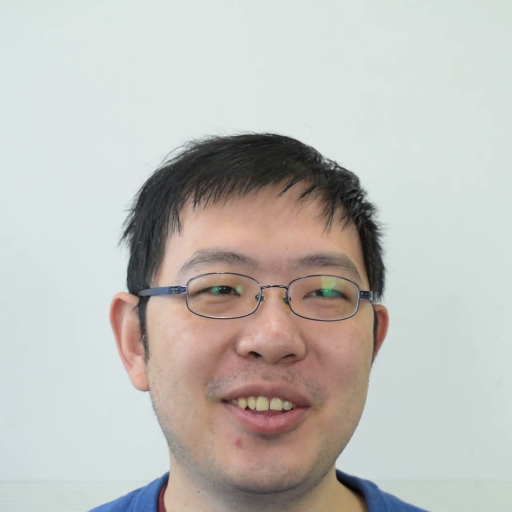}
        }
    \subfloat[Biden]{
        \includegraphics[width =0.1\textwidth]
        {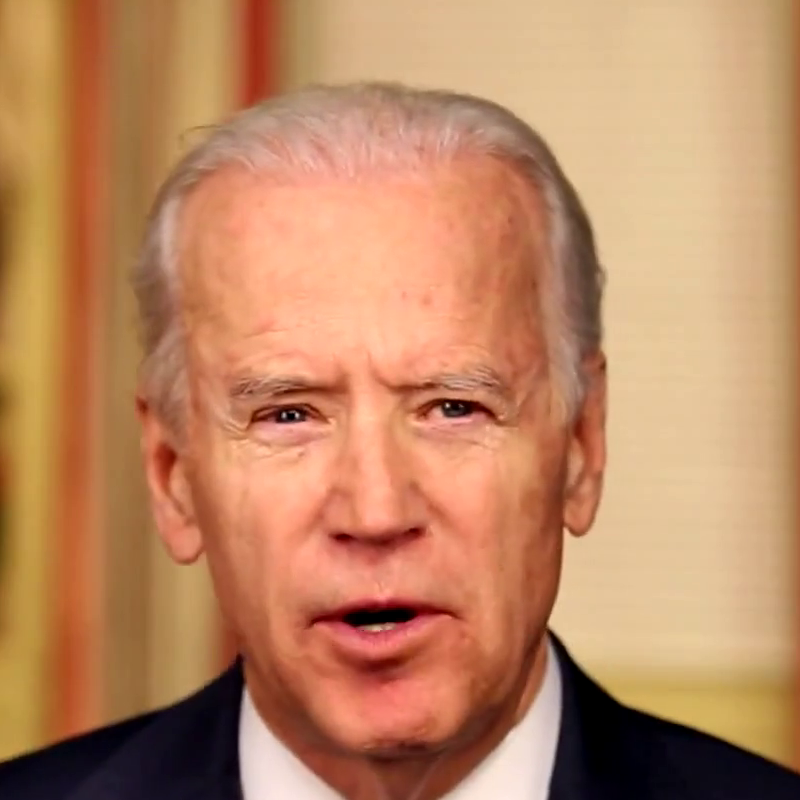}
        } \\
    \subfloat[May]{
        \includegraphics[width =0.1\textwidth]
        {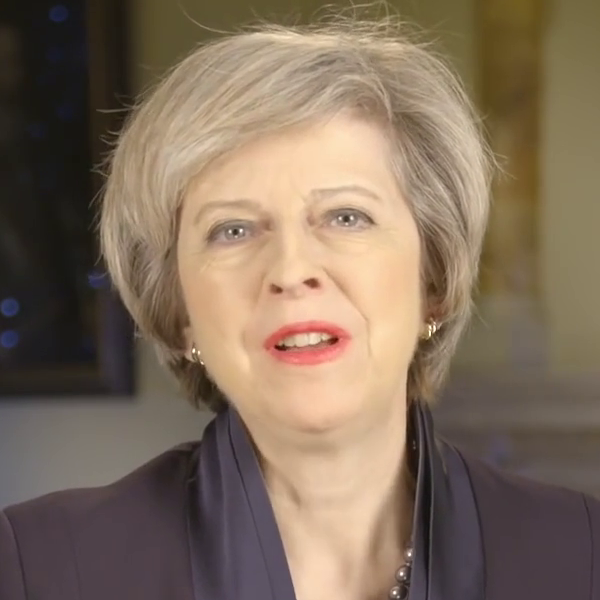}
        }
    \subfloat[Obama]{
        \includegraphics[width =0.1\textwidth]
        {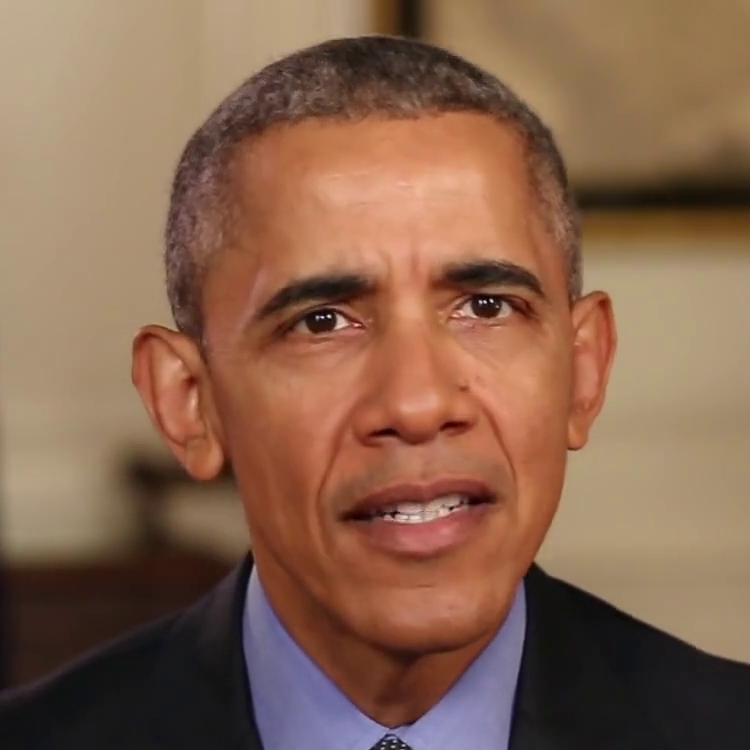}
        }
    \subfloat[Trudeau]{
        \includegraphics[width =0.1\textwidth]
        {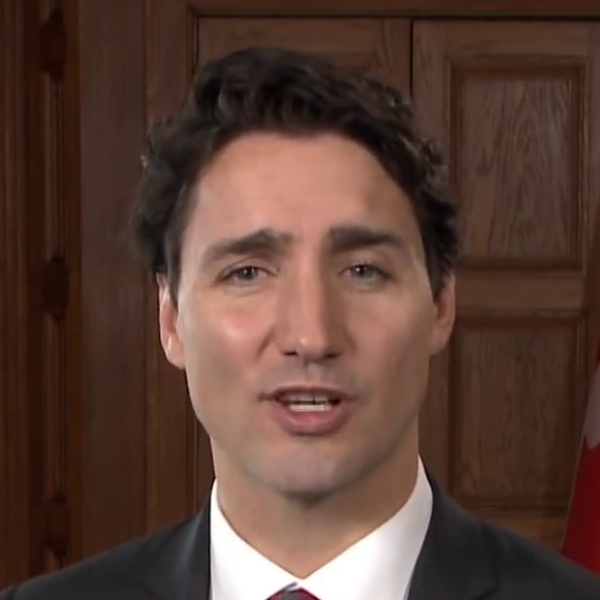}
        }
    \subfloat[Tusk]{
        \includegraphics[width =0.1\textwidth]
        {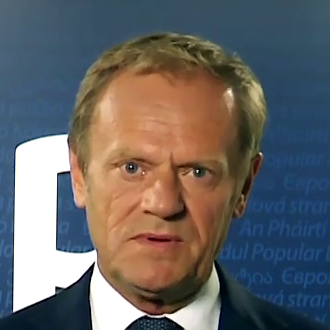}
        }
    \caption{Example images of the dataset used in the experiments. Each image corresponds to a monocular video.}
    \label{img:1.1:dataset_example}
\end{figure}

The datasets used in the experiments are from Head2Head \cite{koujan2020head2head,doukas2021head2head++} and NerFACE \cite{gafni2021dynamic}. The monocular videos used in the experiments are shown in \cref{table:1.1:dataset}. We show the number of frames and the data source corresponding to each video. Some examples are shown in \cref{img:1.1:dataset_example}. Each image corresponds to a monocular video for neural portrait generation. 

For the images used for experiments, we pre-crop the face images in the monocular video, keeping a fixed background and camera viewpoint. To extract the face parameters, we use some advanced face trackers. Specifically, we use DECA \cite{feng2021learning} to extract the pose coefficients $p \in R^6$ and expression coefficients $e \in R^{50}$ of the 3DMM \cite{blanz1999morphable,booth20163d,paysan20093d,li2017learning}. Then we use Gaze360 \cite{kellnhofer2019gaze360} to extract the gaze coefficients $g \in R^2$. In addition, we use DeepSpeech \cite{hannun2014deep,amodei2016deep} to extract audio features $a \in R^{16 \times 29}$ from the input audio.

% We define the monocular videos used to provide personalized prior as \textit{performing data}, the data used to provide input priors as \textit{auxiliary data}, and the data used to test the quality of the generation as \textit{driven data}. We use the labels defined in \cref{table:1.1:dataset} to represent the corresponding data. To ensure consistent data sizes during training, we use about 5k images for performing data and 11k to 16k images for auxiliary data.

% We define the monocular videos used to provide personalized prior as \textit{performing data}, the data used to provide input priors as \textit{auxiliary data}, and the data used to test the quality of the generation as \textit{driven data}. In Table \ref{table:1.1:dataset_spec}, we show the performing data, auxiliary data, and driven data used in each experiment, as well as their corresponding amount of data.

% ****************************************************
\section{Training Detail}
During training, we define the monocular video used to provide personalized prior as \textit{performing data}, and the dataset used to provide morphable prior as \textit{auxiliary data}. Our framework uses the above data to perform a prior-guided two-stage training strategy as follows. 

Taking the video-driven application as an example, in \textit{stage I}, we use performing data to train the generator during reconstruction training. A total of 100,000 iterations are performed in this stage with the batch set to 16, using the Adam \cite{kingma2014adam} optimizer with a learning rate of 0.0005. For the hyperparameters in the loss function, we set $\alpha_1=100$, $\alpha_2=\alpha_3=0$, and $\alpha_4=1$. In \textit{stage II}, we perform scalable training with both performing data and auxiliary data to train the generator and the discriminator. A total of 100,000 iterations are performed in this stage with the batch set to 8. We use the Adam optimizer where the learning rate is 0.0001 for the generator and 0.0004 for the discriminator. We set $\alpha_1=100$ and $\alpha_2=\alpha_3=\alpha_4=1$ for the loss function. For the audio-driven application, the audio-related network is trained with the generator. 

Our method is implemented based on the Pytorch \cite{paszke2019pytorch} framework, and trained on two GeForce RTX 3090. As shown in \cref{img:mlp}, we encode the face parameters using the positional encoding function, where $N_x = 10$ for the coordinate and $N_p = N_g = 4$ for the pose coefficients and gaze coefficients. In addition, the dimension of the learnable latent variable $v$ is set to 32.

\begin{table}[t]
    \centering
    % \resizebox{0.47\textwidth}{!}{
    \begin{tabular}{clcl} 
    \hline
    Label & Portrait & $N$ & Data Source    \\ 
    \hline
     (a) & ID.1        &  5513    &  \multirow{3}{*}{NerFACE}    \\
     (b) & ID.2        &  5518    &    \\
     (c) & ID.3        &  5441    &    \\
    \hline
     (d) & Biden        &  5755     & \multirow{5}{*}{Head2Head}    \\
     (e) & May          &  6050     &     \\
     (f) & Obama        &  2218     &     \\
     (g) & Trudeau      &  5035     &     \\
     (h) & Tusk         &  3650     &     \\
    \hline
    \end{tabular}
    % }
    \caption{Illustration of the dataset used in the experiments. $N$ defines the number of frames in a monocular video.}
    \label{table:1.1:dataset}
\end{table}

% \begin{figure*}[htbp]
%     \centering
%     \subfloat[ID.1]{
%         \includegraphics[width =0.23\textwidth]
%         {img/sup/ID1.png}
%         }
%     \subfloat[ID.2]{
%         \includegraphics[width =0.23\textwidth]
%         {img/sup/ID2.png}
%         }
%     \subfloat[ID.3]{
%         \includegraphics[width =0.23\textwidth]
%         {img/sup/ID3.png}
%         }
%     \subfloat[Biden]{
%         \includegraphics[width =0.23\textwidth]
%         {img/sup/Biden.png}
%         } \\
%     \subfloat[May]{
%         \includegraphics[width =0.23\textwidth]
%         {img/sup/May.png}
%         }
%     \subfloat[Obama]{
%         \includegraphics[width =0.23\textwidth]
%         {img/sup/Obama.png}
%         }
%     \subfloat[Trudeau]{
%         \includegraphics[width =0.23\textwidth]
%         {img/sup/Trudeau.png}
%         }
%     \subfloat[Tusk]{
%         \includegraphics[width =0.23\textwidth]
%         {img/sup/tusk.png}
%         }
%     \caption{Example images of the dataset used in the experiments. Each image corresponds to a monocular video.}
%     \label{img:1.1:dataset_example}
% \end{figure*}

% *********************************************************
\section{Network Structure}
We use an efficient network architecture consisting of a 2D coordinate-based generator, a discriminator, and an optional audio-related network.

\begin{figure*}[t]
    \centering
    \includegraphics[width = 0.97\textwidth]{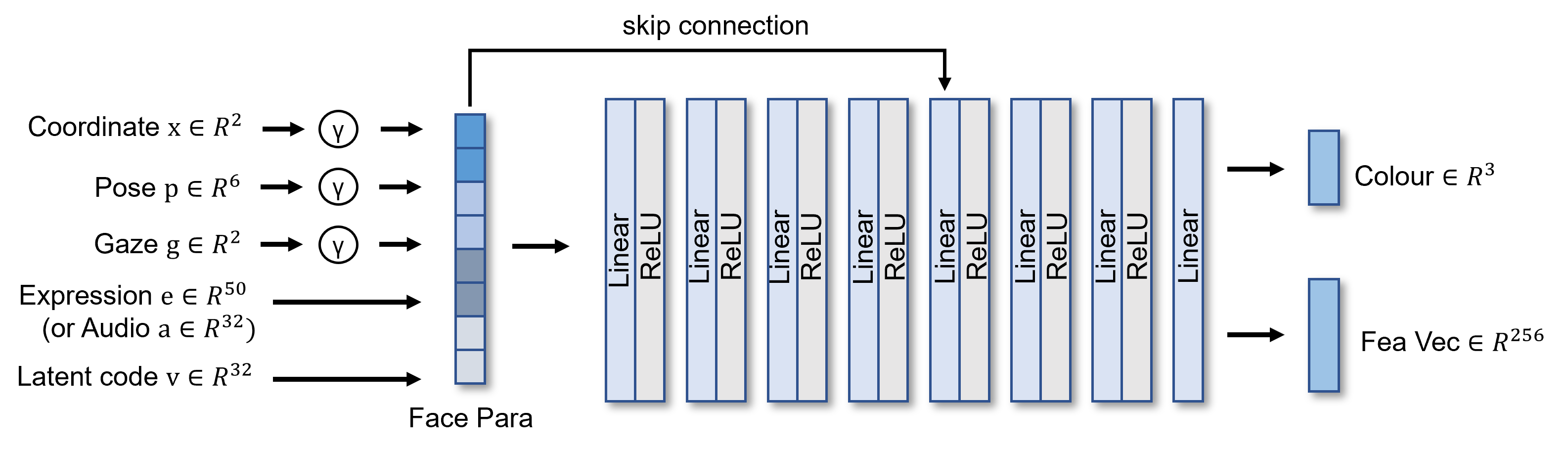}
    \caption{The architecture of the MLP network.}
    \label{img:mlp}
\end{figure*}

\begin{figure*}[t]
    \centering
    \includegraphics[width = 0.97\textwidth]{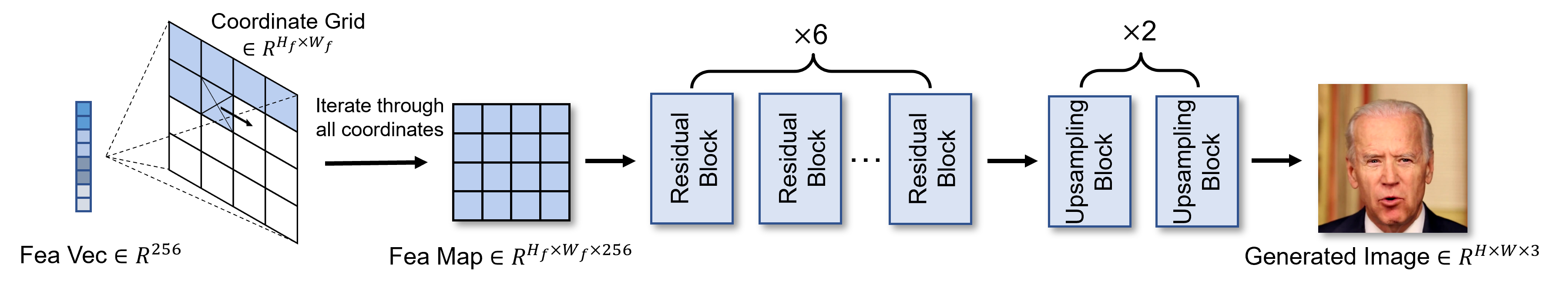}
    \caption{The architecture of the Decoder network.}
    \label{img:decoder}
\end{figure*}

\noindent{\textbf{Generator.}} Our generator network includes a 2D coordinate-based multilayer perceptron (MLP) and a decoder based on convolutional neural networks (CNN). The MLP network consists of eight fully connected layers with all output dimensions of 128. The network structure of the MLP is illustrated in \cref{img:mlp}. In addition to predicting the feature vector based on the input parameter vector, MLP also predicts the pixel value to construct an improved reconstruction loss. For more details, please refer to DNP \cite{doukas2023dynamic}.

The CNN-based decoder contains six residual blocks and two upsampling blocks. Unlike DNP, we use bilinear interpolation rather than transposed convolution for upsampling because we observe that transposed convolution produces checkerboard artifacts \cite{odena2016deconvolution}. By resizing the coordinate grid to $64 \times 64$ or $128 \times 128$, we can generate images with resolutions of $256 \times 256$ or $512 \times 512$, respectively. The network structure of the decoder is illustrated in \cref{img:decoder}. 

\noindent{\textbf{Discriminator.}} We build the discriminator network based on PatchGANs \cite{isola2017image,zhu2017unpaired}. In addition, We use spectral normalization \cite{miyato2018spectral} to make the network stable during training.

\noindent{\textbf{Audio-related network.}} Following AD-NeRF \cite{guo2021ad}, We first use DeepSpeech \cite{hannun2014deep,amodei2016deep} to transform speech into audio features with a window size of 16. Then we use a 1D-convolutional network and a self-attention \cite{zhang2019self} network to extract audio vectors of dimension 32. In the audio-driven task, the audio vector is fed into the generator network like other semantic parameters.
% 

% *********************************************************
% \section{Experiment}
\section{Discussion}

\noindent \textbf{Execution Speed.} Benefiting from the simple network structure, our method enables real-time inference. The evaluation results of the execution speed are presented in \cref{tab:speed-test}. Our method takes only 14.3 milliseconds and 28.1 milliseconds to generate an image with a resolution of $256\times256$ and an image with a resolution of $512\times512$, respectively. The experiment is performed on a GeForce RTX 3090.

\noindent{\textbf{Quality vs. Consistency.}} It should be noted that there are differences in the 3DMM neutral templates corresponding to different identities. Even if the 3DMM coefficients corresponding to the generated images are consistent with the driven data, it is not guaranteed that the expressions and poses of the generated images are exactly identical to those in the driven images. Therefore, instead of excessively pursuing such consistency, our method aligns the training and testing data at the expense of certain face parameter consistency, effectively improving the generation quality. This assumption can be observed in the cross-reenactment experiments of the main text.

\begin{table}[t]
    \centering
    % \resizebox{0.45\textwidth}{!}{
    \begin{tabular}{cccccc} 
    \hline
    \multirow{2}{*}{Method} & \multirow{2}{*}{Param} & \multicolumn{2}{c}{$256\times256$} & \multicolumn{2}{c}{$512\times512$}   \\
     & & Time &Fps &  Time &Fps   \\
    \hline
    {Ours}  &7.6M   & 14.3 &70.2 & 28.1 &35.6   \\ 
    \hline
    \end{tabular}
    % }
    \caption{The execution speed of our method. Time is reported in milliseconds (msec). Please note that all reported numbers refer to the forward pass time of models during inference, without considering data pre-processing.}
    \label{tab:speed-test}
\end{table}

% {
%     \small
%     \bibliographystyle{ieeenat_fullname}
%     \bibliography{main}
% }
% *****************************************************
\end{document}